\documentclass[10pt,journal,compsoc]{IEEEtran}

\usepackage{times}
\usepackage{epsfig}
\usepackage{graphicx}
\usepackage{amsmath}
\usepackage{amssymb}
\usepackage{floatrow}
\usepackage{url}
\usepackage{color}
\usepackage{xcolor}
\usepackage{empheq}
\usepackage{amsfonts}
\usepackage{dsfont}
\usepackage{caption}
\usepackage{enumitem}
\usepackage{multirow}
\usepackage{booktabs}
\usepackage{threeparttable}
\usepackage{wrapfig}
\usepackage{makecell}
\usepackage{tablefootnote}
\usepackage{footnote}
\usepackage{amsthm}

\usepackage{times}
\usepackage{epsfig}
\usepackage{graphicx}
\usepackage{subfigure}
\usepackage{amsmath}
\usepackage{amssymb}
\usepackage{amsthm}

\usepackage{mathrsfs}
\usepackage{hyperref}
\usepackage{color}
\usepackage{xcolor}

\usepackage{algorithm}
\usepackage{algorithmic}
\usepackage{array}
\usepackage{booktabs}
\usepackage{multirow}

\usepackage{caption}
\usepackage{enumitem}
\usepackage{color}

\newcommand{\tabincell}[2]{\begin{tabular}{@{}#1@{}}#2\end{tabular}}
\newcommand{\eg} {\emph{e.g. }}
\newcommand{\ie} {\emph{i.e. }}
\newcommand{\vs} {\emph{v.s. }}
\newcommand{\etal} {\emph{et al. }}
\newcommand*{\tran}{{^{\mkern-1.5mu\mathsf{T}}}}
\newcommand{\w} {{ \mathbf{w} }}
\newcommand{\x} {{ \mathbf{x} }}

\ifCLASSINFOpdf
\else
\fi

\begin{document}
\title{Switchable Normalization for Learning-to-Normalize Deep Representation}


\author{Ping Luo, Ruimao Zhang, Jiamin Ren, Zhanglin Peng, Jingyu Li

\thanks{Ping Luo is with Department of Computer Science, The University of Hong Kong, Hong Kong,
China. (Email: pluo.lhi@gmail.com).}
\thanks{Ruimao Zhang, Jiamin Ren, Zhanglin Peng are with SenseTime Research, Shenzhen, China. (Email: ruimao.zhang@ieee.org; renjiamin@sensetime.com; pengzhanglin@sensetime.com ). Ruimao Zhang is also with The Chinese University of Hong Kong. He is the corresponding author of this paper.}
\thanks{Jingyu Li is with Department of Electronic Engineering, The Chinese University of Hong Kong, Hong Kong,
China. (Email: jingyuli@cuhk.edu.hk ). }
}


\markboth{IEEE TRANSACTIONS ON PATTERN ANALYSIS AND MACHINE INTELLIGENCE}%
{Shell \MakeLowercase{\textit{et al.}}: Bare Demo of IEEEtran.cls for IEEE Transactions on Magnetics Journals}

\IEEEcompsoctitleabstractindextext{
\begin{abstract}
We address a learning-to-normalize problem by proposing Switchable Normalization (SN), which learns to select different normalizers for different normalization layers of a deep neural network.
SN employs three distinct scopes to compute statistics (means and variances) including a channel, a layer, and a minibatch. SN switches between them by learning their importance weights in an end-to-end manner.
It has several good properties.
First,
it adapts to various network architectures and tasks (see Fig.\ref{fig:intro}).
Second, it is robust to a wide range of batch sizes,
maintaining high performance even when small minibatch is presented (\eg 2 images/GPU).
Third, SN does not have sensitive hyper-parameter, unlike group normalization that searches the number of groups as a hyper-parameter.
Without bells and whistles, SN outperforms its counterparts on various challenging benchmarks, such as ImageNet, COCO, CityScapes, ADE20K, MegaFace and Kinetics.
%
Analyses of SN are also presented to answer the following three questions:
(a) Is it useful to allow each normalization layer to select its own normalizer?
(b) What impacts the choices of normalizers?
(c) Do different tasks and datasets prefer different normalizers?
We hope SN will help ease the usage and understand the normalization techniques in deep learning.
The code of SN has been released at \url{https://github.com/switchablenorms}.

\end{abstract}

}

\maketitle
\IEEEpeerreviewmaketitle

\section{Introduction}\label{sec:intro}

Normalization techniques are effective components in deep learning,
advancing many research fields such as natural language processing, computer vision, and machine learning.
In recent years, many normalization methods such as Batch Normalization (BN) \cite{C:BN}, Instance Normalization (IN) \cite{IN}, and Layer Normalization (LN) \cite{LN} have been developed.
%
Despite their great successes, existing practices often employed the same normalizer in all normalization layers of an entire network, rendering suboptimal performance. Also, different normalizers are used to solve different tasks, making model design cumbersome.
%

\begin{figure}[htbp] \centering
\subfigure[] {
\begin{minipage}[c]{0.9\linewidth}
\centering
\includegraphics[width=1\textwidth]{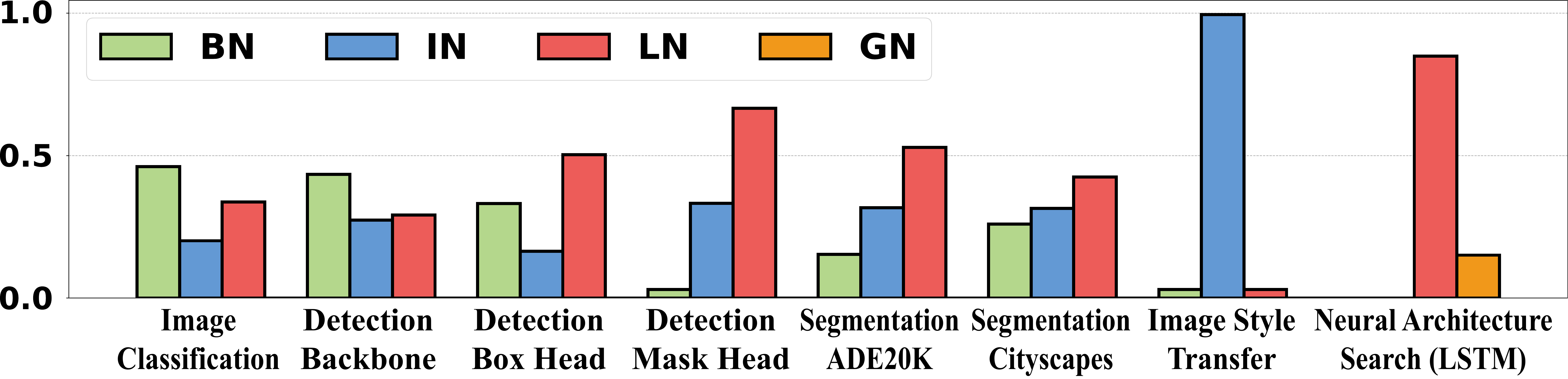}
\end{minipage}%
}%
\hspace{-0.0in}
\subfigure[] {
\begin{minipage}[c]{0.72\linewidth}
\centering
\includegraphics[width=1\textwidth]{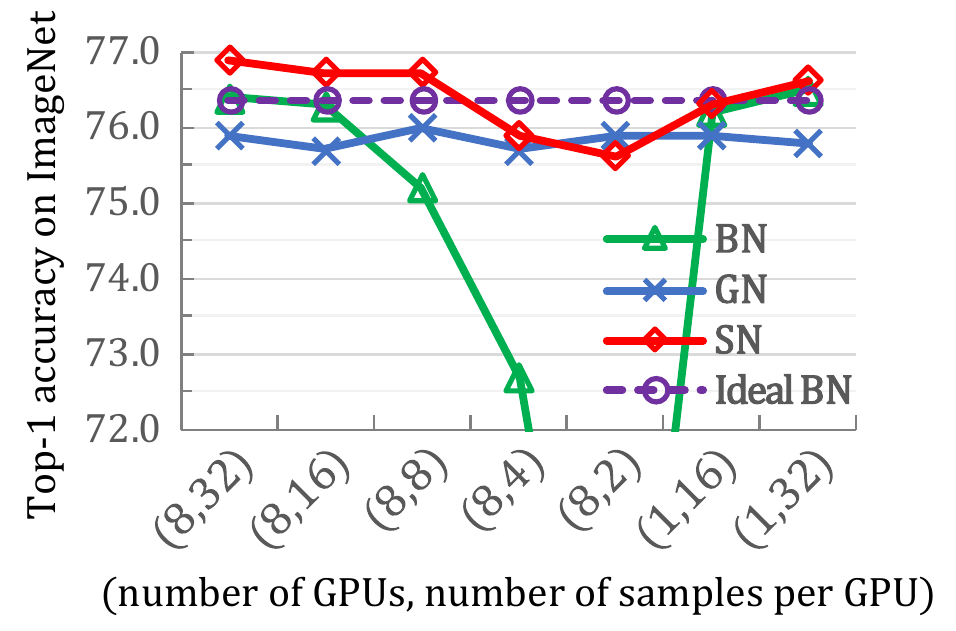}
\end{minipage}%
}%
\vspace{-10pt}
\caption{ \small{{(a)} shows that SN adapts to various networks and tasks by learning importance ratios to select normalizers. In (a), a ratio is between 0 and 1 and all ratios of each task sum to 1.
{(b)} shows the top-1 accuracies of ResNet50 trained with SN on ImageNet and compared with BN and GN in different batch settings.
The gradients in training are averaged over all GPUs and the statistics of normalizers are estimated in each GPU.
For instance, all methods are compared to an ideal case, `ideal BN', whose accuracies are 76.4\% for all settings. This ideal case cannot be obtained in practice.
In fact, when the minibatch size decreases, BN's accuracies drop significantly, while SN and GN both maintain reasonably good performance. SN surpasses or is comparable to both BN and GN in all settings.
}}
\label{fig:intro}
\vspace{-10pt}
\end{figure}

To address the above issues, we propose
\emph{Switchable Normalization (SN)}, which combines three types of statistics estimated
channel-wise, layer-wise, and minibatch-wise by using IN, LN, and BN respectively.
%
SN switches among them by learning their importance weights.
By design, \emph{SN is adaptable to various deep networks and tasks}.
For example, the ratios of IN, LN, and BN in SN are compared in multiple tasks as shown in Fig.\ref{fig:intro} (a).
We see that using one normalization method uniformly is not optimal for these tasks.
For instance, image classification and object detection prefer the combination of three normalizers.
In particular, SN chooses BN more than IN and LN in image classification and the backbone network of object detection, while LN has larger weights in the box and mask heads.
For artistic image style transfer \cite{style}, SN selects IN.
For neural architecture search, SN is applied to LSTM where LN is preferable than group normalization (GN) \cite{GN}, which is a variant of IN by dividing channels into groups.

The selectivity of normalizers makes \emph{SN robust to minibatch size}. As shown in Fig.\ref{fig:intro} (b), when training ResNet50 \cite{C:resnet} on ImageNet \cite{C:ImageNet} with different batch sizes, SN is close to the ``ideal case'' more than BN and GN.
For $(8,32)$ as an example\footnote{In this work, {minibatch size} refers to the number of samples per GPU, and {batch size} is `$\#$GPUs' times `$\#$samples per GPU'. A batch setting is denoted as a 2-tuple, \emph{($\#$GPUs, $\#$samples per GPU)}.}, ResNet50 trained with SN is able to achieve 76.9\% top-1 accuracy, surpassing BN and GN by 0.5\% and 1.0\% respectively.
In general, SN obtains better or comparable results than both BN and GN in all batch settings.

Overall, this work has three key \textbf{contributions}.
{(1)} We introduce Switchable Normalization (SN), which
%
%
%
is applicable in both CNNs and RNNs/LSTMs, and improves the other normalization techniques on many challenging benchmarks and tasks including image recognition in ImageNet \cite{J:ImageNetChallenge}, object detection in COCO \cite{lin2014microsoft}, scene parsing in Cityscapes \cite{C:cityscape} and ADE20K \cite{C:ADE20K}, face recognition in MegaFace~\cite{C:megaface}, video recognition in Kinetics \cite{kay2017kinetics},
artistic image stylization \cite{style} and neural architecture search \cite{ENAS}.
(2) The analyses of SN are presented where multiple normalizers can be compared and understood with geometric interpretation.
In addition, through systematic experiments, we answer three critical questions:
i) Is it useful to allow each normalization layer to select its own normalizer?
ii) What impacts the choices of normalizers?
iii) Do different tasks and datasets prefer different normalizers?
The answer of these questions give the suggestions of how to use SN in modern deep neural networks.
(3)
%
By enabling \emph{each normalization layer in a deep network to have its own operation}, SN helps ease the usage of normalizers, pushes the frontier of normalization in deep learning, as well as opens up new research direction.
We believe that all existing models could be reexamined with this new perspective.
We make the code of SN available and
recommend it as an alternative of existing handcrafted approaches.

In the following sections, we first review the related work in Sec.\ref{sec:related} and then present SN in Sec.\ref{sec:SN1}.
The performance of SN is evaluated extensively in Sec.\ref{sec:exp}.
We conclude our work and summarize some advices that help training CNNs with SN in Sec.~\ref{sec:conclusion}. The future research directions are also presented in this section.


\section{Related Work}\label{sec:related}

\begin{table*}[t]
 \centering
 \begin{tabular}{l|ccc|ccc
 }
\hline
& \multicolumn{3}{c|}{Parameter} &\multicolumn{3}{c}{Statistical Estimation}
\\
\cline{2-7} & \tabincell{c}{params}
  & \tabincell{c}{$\#$params}& \tabincell{c}{hyper-\\params} & \tabincell{c}{statistics}& \tabincell{c}{computation\\complexity} & \tabincell{c}{$\#$statistics}
   %
 \\
\hline
BN \cite{C:BN}& $\gamma,\beta$ & $2C$& $p,\epsilon$ & $\mu,\sigma,\mu^\prime,\sigma^\prime$ & $\mathcal{O}(NCHW)$& $2C$ 
\\
IN \cite{IN}&$\gamma,\beta$ &$2C$ & $\epsilon$ & $\mu,\sigma$ &$\mathcal{O}(NCHW)$ & $2CN$ \\
LN \cite{LN}&$\gamma,\beta$ & $2C$& $\epsilon$ & $\mu,\sigma$ &$\mathcal{O}(NCHW)$& $2N$ \\
GN \cite{GN}& $\gamma,\beta$ &$2C$& $g,\epsilon$ &$\mu,\sigma$ &$\mathcal{O}(NCHW)$& $2gN$ \\
\hline
BRN \cite{BRN} &$\gamma,\beta$ &$2C$& $p,\epsilon,r,d$ & $\mu,\sigma,\mu^\prime,\sigma^\prime$ &$\mathcal{O}(NCHW)$& $2C$ \\
BKN \cite{BKN} & $A$ &$C^2$& $p,\epsilon$ & $\mu,\Sigma,\mu^\prime,\Sigma^\prime$ &$\mathcal{O}(NC^2HW)$ & $C+C^2$ \\
WN \cite{WN} &$\gamma$ &$C$& -- & -- &-- & -- \\
\hline
SN & \tabincell{c}{$\gamma,\beta$,\\$\{w_k\}_{k\in\Omega}$} & $2C+6$& $\epsilon$ & $\{\mu_k,\sigma_k\}_{k\in\Omega}$ &$\mathcal{O}(NCHW)$& \tabincell{c}{$2C+2N$\\$+2CN$} \\
 \hline
 \end{tabular}
 \caption{\small{\textbf{Comparisons of normalization methods}.
 First, we compare
 their types of parameters, numbers of parameters ($\#$params), and hyper-parameters.
 Second, we compare types of statistics, computational complexity to estimate statistics, and numbers of statistics ($\#$statistics).
 %
 %
 %
 Specifically,
 $\gamma,\beta$ denote the scale and shift parameters.
 $\mu,\sigma,\Sigma$ are a vector of means, a vector of standard deviations, and a covariance matrix.
 $\mu^\prime$ represents the moving average.
 Moreover, $p$ is the momentum of moving average, $g$ in GN is the number of groups, $\epsilon$ is a small value for numerical stability, and $r,d$ are used in BRN.
 In SN, $k\in\Omega$ indicates a set of different kinds of statistics, $\Omega=\{\mathrm{in,ln,bn}\}$, and $w_k$ is an importance weight of each kind.
 }}\label{tab:comp_method}
\end{table*}

\textbf{Normalization.}
As one of the most significant component in deep neural networks, normalization technique has achieved much attention in literature~\cite{C:BN,IN,LN,GN,WN,C:SpectralNorm}.
According to the normalized space, related methods can be divided into two groups:
methods normalizing activated representation over feature space such as~\cite{C:BN,IN,LN,GN},
and methods normalizing weights over the parameter space like~\cite{WN,C:SpectralNorm}.

Among former group, Batch Normalization (BN)~\cite{C:BN} is one of the most well-known method,
which is motivated by the fact that whitening input features (\ie centering, decorrelating and scaling)~\cite{WNN,GWNN} of each hidden layer can mimic the fast convergence of natural gradient descent (NGD)~\cite{B:InformationGeometry} by using stochastic gradient descent (SGD).
Even though BN only pursuit standardization (\ie centering and scaling) instead of whitening transformation, it still stabilizes the training process and boost the performance of neural network.
To better understand the effectiveness of BN, a lot of theory studies have been proposed~\cite{A:Shattered,C:HowNorm,C:NormMatter,regBN,C:MFTNnorm,C:Rate-Tuning}.
For example, Santurkar \textit{et. al.}~\cite{C:HowNorm} argued that
BN did not help covariate shift but contributed to smooth loss landscape.
Hoffer \textit{et. al.}~\cite{C:NormMatter} developed a framework to decouple weights' norm from the underlying optimized objective.
Luo \textit{et. al.}~\cite{regBN} decomposed BN into population normalization (PN)
and gamma decay as an explicit regularization, and claimed that BN allowed bigger learning rate in the training process.
The similar conclusion was also achieved by Bjorck\textit{et. al.}~\cite{C:Understading}.
In~\cite{C:MFTNnorm}, Yang \textit{et. al.} developed a mean field theory for batch normalization in fully-connected feedforward neural networks.
Arora \textit{et. al.}~\cite{C:Rate-Tuning} found that BN had the ability to allow gradient descent to succeed with less tuning of learning rates.

Despite a series of theory analyses, a lot of work have been proposed in practice to address the issue that BN required reasonable mini-batch size to estimate the mean and variance.
Ba \textit{et. al.}~\cite{LN} proposed Layer Normalization (LN) to calculate the mean and variance for each sample on a single layer.
By leveraging BN and LN, Ren~\textit{et. al.}~\cite{Normalizing} proposed Division Normalization to explore spatial region.
In~\cite{IN}, Instance Normalization (IN) is proposed as a channel-wise normalization method to filter out complex appearance variance~\cite{C:IBN}.
Wu and He~\cite{GN} further proposed Group Normalization (GN) by dividing the channels into groups and computing within each group the mean and variance for normalization.
Luo~\textit{et. al.}~\cite{C:DN} proposed Dynamic Normalization (DN) to learn arbitrary normalization for different convolutional layers.
Such method achieved stable accuracy in a wide range of batch sizes.
Other attempts to deal with instability of BN with small batch-size including Batch Renormalization (BRN)~\cite{BRN}, Batch Kalman Normalization (BKN)~\cite{BKN} and Stream Normalization (StN)~\cite{C:StreamNrom}.

Some work also proposed to promote the performance of BN on other aspects.
For example, Huang~\textit{et. al.}~\cite{DBN} proposed Decorrelated Batch Normalization (DBN) to whiten the activations of each layer within a mini-batch to improve optimization efficiency and generalization ability of BN.
In~\cite{C:IBN}, Pan~\textit{et. al.} showed that the hybrid of BN and IN in the neural networks can greatly strengthen the generalization ability of CNN on different domains.
In~\cite{ConBN}, Perez~\textit{et. al.} adopted Conditional Batch Normalization (CBN) to affect the feature representation of each sample independently in BN layers, and achieved promising results on visual question answering task.
In~\cite{A:SW}, Pan~\textit{et. al.}  proposed Switchable Whitening (SW) to select different whitening methods as well as standardization methods in a single layer.

For the methods normalizing the weights, Salimans~\textit{et. al.}~\cite{WN} firstly proposed Weight Normalization (WN) to normalize the weight for each neuron to decouple the length of weight vectors from their directions.
Huang~\textit{et. al.}，\cite{C:CWN} further introduced Centered Weight Normalization (CWN) to further powered WN by centering the input weight.
In recent studies on Generative Adversarial Network (GAN)~\cite{C:GAN}, Miyato~\textit{et. al.}~\cite{C:SpectralNorm} proposed Spectral Normalization (SpN) to re-scale the weight vector by dividing its largest singular value to satisfy the Lipschitz constraint.
This method stabilized the training process of the discriminator in GAN, and effectively improved the quality of generated images.


In Table \ref{tab:comp_method},
we compare SN with five popular normalization methods, \ie BN, IN, LN, GN and WN, as well as three variants of BN including Batch Renormalization (BRN) and Batch Kalman Normalization (BKN) in details.
In general,
we see that SN possesses comparable numbers of parameters and computations, as well as rich statistics.
\textbf{First}, although SN has richer statistics, the computational complexity to estimate them is comparable to previous methods, as shown in the second portion of Table \ref{tab:comp_method}.
As introduced in Sec.\ref{sec:SN1}, IN, LN, and BN estimate the means and variances along axes $(H,W)$, $(C,H,W)$, and $(N,H,W)$ respectively, leading to $2CN$, $2N$, and $2C$ numbers of statistics.
Therefore, SN has $2CN+2N+2C$ statistics by combining them.
Although BKN has the largest number of $C+C^2$ statistics,
it also has the highest computations because it estimates the covariance matrix other than the variance vector. Also, approximating the covariance matrix in a minibatch is nontrivial as discussed in \cite{WNN,GWNN,EigenNet}.
BN, BRN, and BKN also compute moving averages.
\textbf{Second}, SN is demonstrated in various networks, tasks, and datasets.
Its applications are much wider than existing normalizers and it also has rich theoretical value that is worth exploring.

\textbf{Meta Learning.}
Our work is also related with meta learning (ML) problem \cite{bilevel,hyperparameter,ENAS},
which can be also used to learn the control parameters in SN. In general, ML is defined as $\min_{\Theta}\mathcal{L}(\Theta,\Phi^\ast)+\min_{\Phi}\varphi\mathcal{L}(\Theta^\ast,\Phi)$ where $\varphi$ is a constant multiplier.
%
Unlike SN trained with a single stage, this loss function is minimized by performing two feed-forward stages iteratively until converged.
First, by fixing the current estimated control parameters $\Phi^\ast$, the network parameters are optimized by $\Theta^\ast=\min_\Theta\mathcal{L}(\Theta,\Phi^\ast)$.
Second, by fixing $\Theta^\ast$, the control parameters are found by $\Phi^\ast=\min_\Phi\varphi\mathcal{L}(\Theta^*,\Phi)$.

The above two stages are usually optimized by using two different sets of training data. 
For example, previous work \cite{DARTS,ENAS} used $\Phi$ to search network architectures from a set of modules with different numbers of parameters and computational complexities. They divided an entire training set into a training and a validation set without overlapping, where $\Phi$ is learned from the validation set while $\Theta$ is learned from the training set.
This is because $\Phi$ would choose the module with large complexity to overfit training data, if both $\Theta$ and $\Phi$ are optimized in the same dataset.

The above 2-stage training increases computations and runtime.
In contrast, $\Theta$ and $\Phi$ for SN can be generally optimized within a single stage in the same dataset, because $\Phi$ regularizes training by choosing different normalizers from $\Omega$ to prevent overfitting.

\section{Switchable Normalization (SN)}\label{sec:SN1}

We describe a general formulation of a normalization layer and then present SN in this section.

\textbf{A General Form.} We take CNN as an illustrative example.
Let $h$ be the input data of an arbitrary normalization layer represented by a 4D tensor $(N,C,H,W)$, indicating number of samples, number of channels, height and width of a channel respectively.
%
Let $h_{ncij}$ and $\hat{h}_{ncij}$ be a pixel before and after normalization, where $n\in[1,N]$, $c\in[1,C]$, $i\in[1,H]$, and $j\in[1,W]$.
Let $\mu$ and $\sigma$ be a mean and a standard deviation.
We have
\begin{equation}
\hat{h}_{ncij}=\gamma\frac{h_{ncij}-\mu}{\sqrt{\sigma^2+\epsilon}}+\beta,\label{eq:normalize}
\end{equation}
where $\gamma$ and $\beta$ are a scale and a shift parameter respectively. $\epsilon$ is a small constant to preserve numerical stability.
Eqn.(\ref{eq:normalize}) shows that each pixel is normalized by using $\mu$ and $\sigma$, and then re-scale and re-shift by $\gamma$ and $\beta$.
%

In practice, IN, LN, and BN share the formulation of Eqn.(\ref{eq:normalize}), but they use different sets of pixels to estimate $\mu$ and $\sigma$. In other words, the numbers of their estimated statistics are different.
In general, we have
\begin{equation}\label{eq:mu_sigma}
\begin{split}
&\mu_k=\frac{1}{|I_k|}\sum_{ {(n,c,i,j)}\in I_k}h_{ncij}, \\
&\sigma_k^2={\frac{1}{|I_k|}\sum_{ {(n,c,i,j)}\in I_k}(h_{ncij}-\mu_k)^2},
\end{split}
\end{equation}
where $k\in\{\mathrm{in},\mathrm{ln},\mathrm{bn}\}$ is used to distinguish different methods.
$I_k$ is a set pixels and $|I_k|$ denotes the number of pixels.
Specifically, $I_{\mathrm{in}}$, $I_{\mathrm{ln}}$, and $I_{\mathrm{bn}}$ are the sets of pixels used to compute statistics in different approaches.

IN was established in the task of artistic image style transfer \cite{style,AIN,C:improvedTexture}. In IN, we have $\mu_{\mathrm{in}},\sigma_{\mathrm{in}}^2 \in \mathbb{R}^{N\times C}$ and $I_{\mathrm{in}}=\{(i,j)|i\in[1,H],j\in[1,W]\}$, meaning that IN has $2NC$ elements of statistics, where each mean and variance value is computed along $(H,W)$ for each channel of each sample.
%

LN \cite{LN} was proposed to ease optimization of recurrent neural networks (RNNs).
In LN, we have $\mu_{\mathrm{ln}},\sigma_{\mathrm{ln}}^2\in\mathbb{R}^{N\times1}$ and $I_{\mathrm{ln}}=\{(c,i,j)|c\in[1,C],i\in[1,H],j\in[1,W]\}$, implying that LN has $2N$ statistical values, where a mean value and a variance value are computed in $(C,H,W)$ for each one of the $N$ samples.

BN \cite{C:BN} was first demonstrated in the task of image classification \cite{C:resnet,C:Alexnet} by normalizing the hidden feature maps of CNNs. In BN, we have $\mu_{\mathrm{bn}},\sigma_{\mathrm{bn}}^2\in\mathbb{R}^{C\times 1}$ and $I_{\mathrm{bn}}=\{(n,i,j)|n\in[1,N],i\in[1,H],j\in[1,W]\}$, in the sense that BN treats each channel independently like IN, but not only normalizes across $(H,W)$, but also the $N$ samples in a minibatch, leading to $2C$ elements of statistics.

\subsection{Formulation of SN}\label{sec:SN}

\begin{figure*}[t]
\centering
\includegraphics[width=0.95\linewidth]{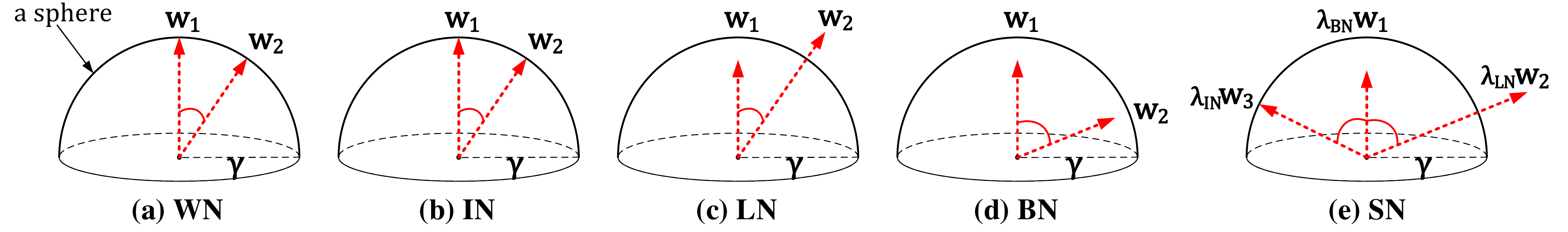}
\caption{\small{Geometric view of directions and lengths of the filters in WN, IN, BN, LN, and SN.
These normalizers are compared in an unify way by represent them by using WN that decomposes optimization of filters into their directions and lengths.
In this way, IN is identical to WN that sets the filter norm to `1' (\ie $\|\w_1\|=\|\w_2\|=1$) and then rescales them to $\gamma$.
LN is less constrained than IN and WN to increase learning ability.
BN increases angle between filters and reduces filter length to improve generalization.
SN inherits all their benefits by learning their importance ratios.
}}\label{fig:geometry}
\end{figure*}

SN has an intuitive expression
\begin{equation}\label{eq:SN}
\hat{h}_{ncij}=\gamma\frac{h_{ncij}-\Sigma_{k\in\Omega} w_k\mu_k}
{\sqrt{\Sigma_{k\in\Omega}w_k^\prime\sigma_k^2+\epsilon}}+\beta,
\end{equation}
where $\Omega$ is a set of statistics estimated in different ways.
In this work, we define $\Omega=\{\mathrm{in},\mathrm{ln},\mathrm{bn}\}$ the same as above where
$\mu_k$ and $\sigma_k^2$ can be calculated by following Eqn.(\ref{eq:mu_sigma}). However, this strategy leads to large redundant computations. In fact, the three kinds of statistics of SN depend on each other.
Therefore we could reduce redundancy by reusing computations,
%
%
%
\begin{eqnarray}\label{eq:reuse}
\mu_{\mathrm{in}}&=&\frac{1}{HW}\sum_{i,j}^{H,W}h_{ncij},
~~~\sigma^2_{\mathrm{in}}=\frac{1}{HW}\sum_{i,j}^{H,W}(h_{ncij}-\mu_{\mathrm{in}})^2,\nonumber\\
\mu_{\mathrm{ln}}&=&\frac{1}{C}\sum_{c=1}^C\mu_{\mathrm{in}},~~~\sigma^2_{\mathrm{ln}}=
\frac{1}{C}\sum_{c=1}^C(\sigma^2_{\mathrm{in}}+\mu_{\mathrm{in}}^2)-\mu_{\mathrm{ln}}^2,\nonumber\\
\mu_{\mathrm{bn}}&=&\frac{1}{N}\sum_{n=1}^N\mu_{\mathrm{in}},~~~\sigma^2_{\mathrm{bn}}=
\frac{1}{N}\sum_{n=1}^N(\sigma^2_{\mathrm{in}}+\mu_{\mathrm{in}}^2)-\mu_{\mathrm{bn}}^2,
\end{eqnarray}
showing that the means and variances of LN and BN can be computed based on IN.
Using Eqn.(\ref{eq:reuse}), the computational complexity of SN is $\mathcal{O}(NCHW)$, which is comparable to previous work.

Furthermore, $w_k$ and $w_k^\prime$ in Eqn.(\ref{eq:SN}) are importance ratios used to weighted average the means and variances respectively.
Each $w_k$ or $w_k^\prime$ is a scalar variable, which is shared across all channels. There are $3\times2=6$ importance weights in SN.
We have $\Sigma_{k\in\Omega}w_k=1$, $\Sigma_{k\in\Omega}w_k^\prime=1$, and $\forall w_k,w_k^\prime\in[0,1]$, and define
%
\begin{equation}\label{eq:SN_w}
w_k=\frac{e^{\lambda_k}}{\Sigma_{z\in\{\mathrm{in},\mathrm{ln},\mathrm{bn}\}}e^{\lambda_z}}~
~\mathrm{and}~~k\in\{\mathrm{in},\mathrm{ln},\mathrm{bn}\}.
\end{equation}
Here each $w_k$ is computed by using a softmax function with
$\lambda_\mathrm{in}$, $\lambda_\mathrm{ln}$, and $\lambda_\mathrm{bn}$ as the control parameters, which can be learned by back-propagation (BP). $w_k^\prime$ are defined similarly by using another three control parameters $\lambda_\mathrm{in}^\prime$, $\lambda_\mathrm{ln}^\prime$, and $\lambda_\mathrm{bn}^\prime$.

\textbf{Training.}
Let $\Theta$ be a set of network parameters (\eg filters) and ${\Phi}$ be a set of control parameters that control the network architecture.
In SN, we have ${\Phi}=\{\lambda_\mathrm{in},\lambda_\mathrm{ln},\lambda_\mathrm{bn},
\lambda_\mathrm{in}^\prime,\lambda_\mathrm{ln}^\prime,\lambda_\mathrm{bn}^\prime\}$.
%
Training a deep network with SN is to minimize a loss function
$\min_{\{\Theta,\Phi\}} \frac{1}{N}\sum_{j=1}^N \mathcal{L}(\mathbf{y}_j,f(\mathbf{x}_j;\Theta,\Phi))$,
where $\{\mathbf{x}_j,\mathbf{y}_j\}_{j=1}^N$ indicates a set of training samples and their labels.
$f(\mathbf{x}_j;\Theta)$ is a function learned by the CNN to predict $\mathbf{y}_j$.
$\Theta$ and $\Phi$ are the parameters of the model that can be optimized jointly by back-propagation (BP).
This training procedure is different from previous meta-learning algorithms
such as network architecture search \cite{bilevel,DARTS,ENAS}.
%
In previous work, $\Phi$ represents as a set of network modules with different learning capacities, where $\Theta$ and $\Phi$ were optimized in two BP stages iteratively by using two training sets that are non-overlapped. For example, previous work divided an entire training set into a training and a validation set. However, if $\Theta$ and $\Phi$ in previous work are optimized in the same set of training data,
$\Phi$ would choose the module with large complexity to overfit these data.
In contrast, SN essentially prevents overfitting by choosing normalizers to improve both learning and generalization ability as discussed below.

\textbf{Implementation.}
SN can be easily implemented in existing softwares such as PyTorch and TensorFlow.
%
The backward computation of SN can be obtained by automatic differentiation (AD) in these softwares.
Without AD, we need to implement back-propagation (BP) of SN, where the errors are propagated through $\mu_k$ and $\sigma_k^2$. We provide the derivations of BP in Appendix.

\subsection{Analyses of SN}

\textbf{Geometric View of SN.}
To understand SN, we theoretically compare SN with BN, IN, and LN by representing them using weight normalization (WN) \cite{WN} that is independent of mean and variance.
Specifically, the computation of WN is defined by $\hat{h}_{\mathrm{wn}}=\frac{h}{\|\w_i\|_2}=\frac{\w_i \tran \x}{\|\w_i\|_2}$, where $\w$ and $\x$ represent a filter and an image patch.
We use $i$ to indicate a filter of the $i$-th channel.
As shown in Fig.\ref{fig:geometry}~(a), WN normalizes the norm of each filter to a unit sphere with length `1', and then rescales the length to $\gamma$ that is a learnable scale parameter. To simplify our discussions, we suppose this scale parameter is shared among all channels.
In other words, WN decomposes the optimization of a filter into its direction and length.


By assuming $\mu=0$ and $\sigma=1$ for all the normalizers, we see that they can be also represented by using WN.
For instance, IN turns into $\hat{h}_{\mathrm{in}}=\frac{h-\w_i\tran\mathds{E}[\x]}{\sqrt{\w_i\tran(\mathds{E}[\x\x\tran]-\mathds{E}[\x]\tran\mathds{E}[\x])\w_i}} = \frac{\w_i \tran \x}{\|\w_i\|_2}$ that is identical to WN as shown in Fig.\ref{fig:geometry}~(b).
For LN, each channel is normalized by all the channels and thus $\hat{h}_{\mathrm{ln}}=\frac{\w_i\tran\x}{\sqrt{\frac{1}{C}\sum_{i=1}^C\|\w_i\|_2^2}}$ where $C$ is the number of channels.
The filter norm in LN is less constrained than WN and IN as visualized in Fig.\ref{fig:geometry}~(c), where the filter norm can be either longer or shorter than $\gamma$, in the sense that LN increases learning capacity of the networks by diminishing regularization.

Furthermore, BN can be represented by WN with a similar formula as IN, where $\x$ indicates the image patch sampled from all of the instances in a mini-batch.
In~\cite{regBN}, Luo \etal showed that BN imposes regularization on norms of all the filters and reduces correlations between filters.
%
Its geometry interpretation can be viewed in Fig.\ref{fig:geometry}~(d), where the filter norms would be shorter and angle between filters would be larger than the other normalizers.
In conclusion, BN improves generalization \cite{C:BN}. In general, we would have the following relationships
\vspace{-5pt}
\begin{eqnarray}\label{eq:order}
&\mathrm{learning~capacity}: LN>BN>IN,\nonumber \\
&\mathrm{generalization~capacity}: BN>IN>LN. \nonumber
\end{eqnarray}
\vspace{-15pt}

Finally, $\hat{h}_{\mathrm{sn}}$ in SN inherits the benefits from all of them and enables balance between learning and generalization ability.
For example, when the batch size is small, the random noise from the batch statistics of BN would be too strong. SN is able to maintain performance by decreasing $w_{\mathrm{bn}}$ and increasing $w_{\mathrm{ln}}$,
such that the regularization from BN is reduced and the learning ability is enhanced by LN.
This phenomenon is supported by our experiment.
%

\begin{table*}[t]
\centering
\begin{tabular}
{l|p{22pt}<{\centering}p{22pt}<{\centering}p{22pt}<{\centering}p{22pt}<{\centering}p{22pt}<{\centering}
|p{22pt}<{\centering}p{22pt}<{\centering}|p{22pt}<{\centering}p{22pt}<{\centering}}
  \hline
  & (8,32) & (8,16) & (8,8)&(8,4) & (8,2)&(1,16)& (1,32)& (8,1)& (1,8)\\
  \hline
  BN
  &76.4 & 76.3 & 75.2 & 72.7 & 65.3 &  76.2 & 76.5& -- &  75.4\\
  GN
  &75.9 & 75.8 & 76.0 & 75.8 & \textbf{75.9} & 75.9 &75.8 & \textbf{75.5} & 75.5\\
  SN
  & \textbf{76.9} & \textbf{76.7} & \textbf{76.7} &  \textbf{75.9} & 75.6 & \textbf{76.3}& \textbf{76.6} & 75.0& \textbf{75.9} \\
  \hline
  {GN}$-${BN} & -0.5 & -0.5 & 0.8 & 3.1 & {10.6} & -0.3 & -0.7& --& 0.1\\
  {SN}$-${BN} 
  & {0.5} & {0.4} & {1.5} & 3.2 &10.3& 0.1 &0.1& -- & 0.5\\
  {SN}$-${GN} & {1.0} & {0.9}& {0.7} & 0.1 &-0.3& 0.4 &0.8& -0.5& 0.4\\
  \hline
 \end{tabular}
\caption{\small{\textbf{Comparisons of top-1 accuracies} on the validation set of ImageNet, by using ResNet50 trained with SN, BN, and GN in different batch size settings. The bracket $(\cdot,\cdot)$ denotes ($\#$GPUs, $\#$samples per GPU). In the bottom part, `GN-BN' indicates the difference between the accuracies of GN and BN. The `-' in $(8,1)$ indicates BN does not converge. The best-performing result of each setting is shown in bold. 
}}\label{tab:comp_SN_1}
\end{table*}

\textbf{Variants of SN.}
SN has many extensions. For instance, a pretrained network with SN can be finetuned by applying the $\mathrm{argmax}$ function on its control parameters where each normalization layer selects only one normalizer, leading to sparse SN. For $(8,32)$ as an example, SN with sparsity achieves top-1 accuracy of 77.0\% in ImageNet with ResNet50, which is comparable to 76.9\% of SN without sparsity.
Moreover, when the channels are divided into groups, each group could select its own normalizer to
%
increase representation power of SN. Our preliminary results suggest that group SN performs better than SN in some senses. For instance, group SN with only two groups boosts the top-1 accuracy of ResNet50 to 77.2\% in ImageNet. The above two variants will be presented as future work due to the length of paper. This work focuses on SN where the importance weights are tied between channels.

\textbf{Inference.} When applying SN in test, the statistics of IN and LN are computed independently for each sample, while BN uses {batch average} \emph{after} training without computing moving average in each iteration.
%
Here {batch average} is performed in two steps. First, we freeze the parameters of the network and all the SN layers, and feed the network with a certain number of mini-batches randomly chosen from the training set.
Second, we average the means and variances produced by all these mini-batches in each SN layer. %
The averaged statistics are used by BN in SN.

We find that the batch average can also achieve good performance by using a small amount of samples to calculate the statistics.
For example, top-1 accuracies of ResNet50 on ImageNet by using batch average with 50k and all training samples are 76.90\% and 76.92\%. They are trained more stable in the early stage and slightly better than 76.89\% of the moving average.
%
%


\section{Experiments}\label{sec:exp}

This section presents the main results of SN in multiple challenging problems and benchmarks, such as ImageNet \cite{J:ImageNetChallenge}, COCO \cite{lin2014microsoft}, Cityscapes \cite{C:cityscape}, ADE20K \cite{C:ADE20K}, MegaFace~\cite{C:megaface}and Kinetics \cite{kay2017kinetics}, where the effectiveness of SN is demonstrated by comparing with existing normalization techniques.
%

%

\begin{figure*}[htbp] \centering
\subfigure[validation curves of SN in different batch sizes.] { 
\begin{minipage}[c]{0.32\linewidth}
\centering
\includegraphics[height=1.6in]{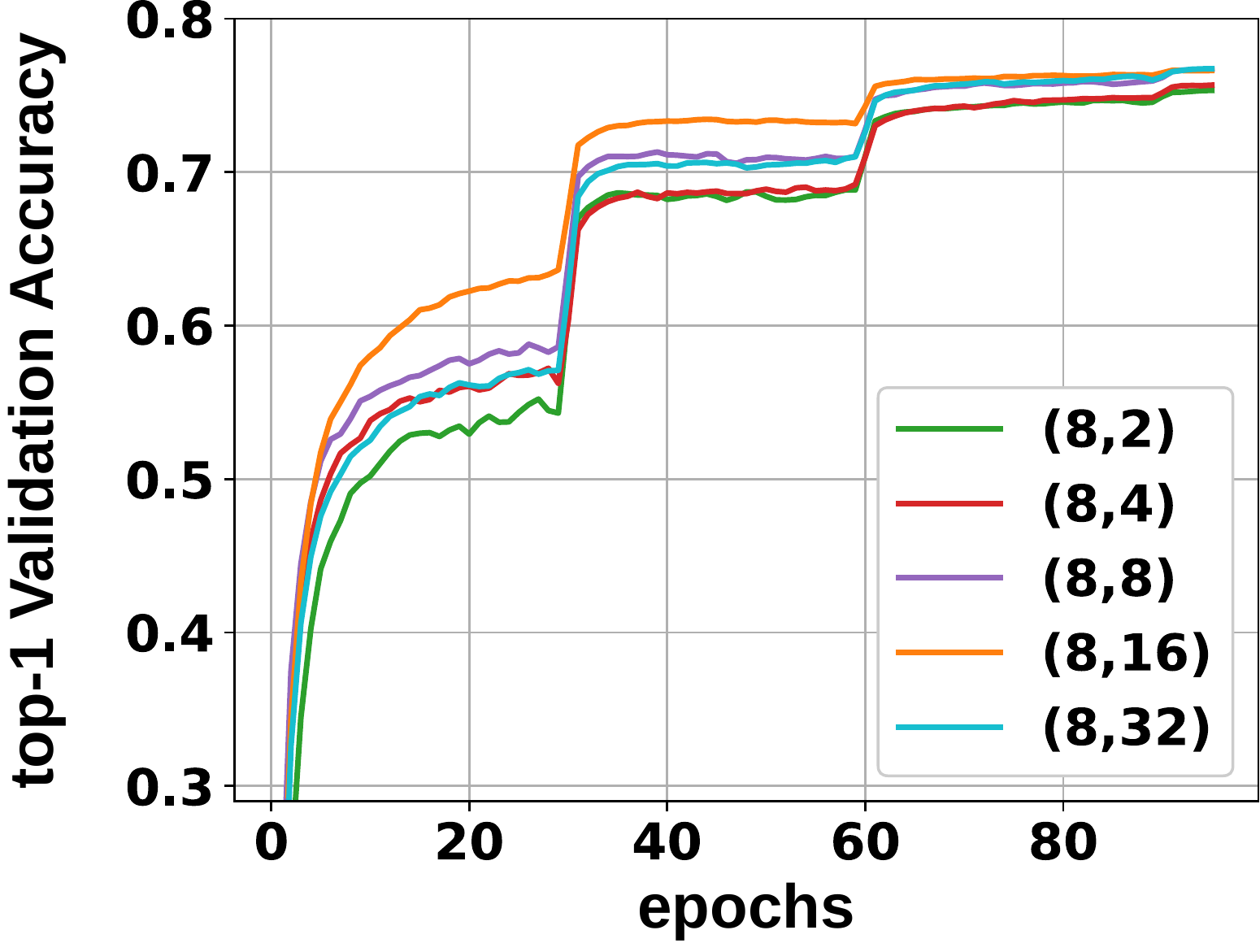}
\end{minipage}%
}%
\hspace{-0.0in}
\subfigure[train and validation curves of (8,32).] { 
\begin{minipage}[c]{0.32\linewidth}
\centering
\includegraphics[height=1.6in]{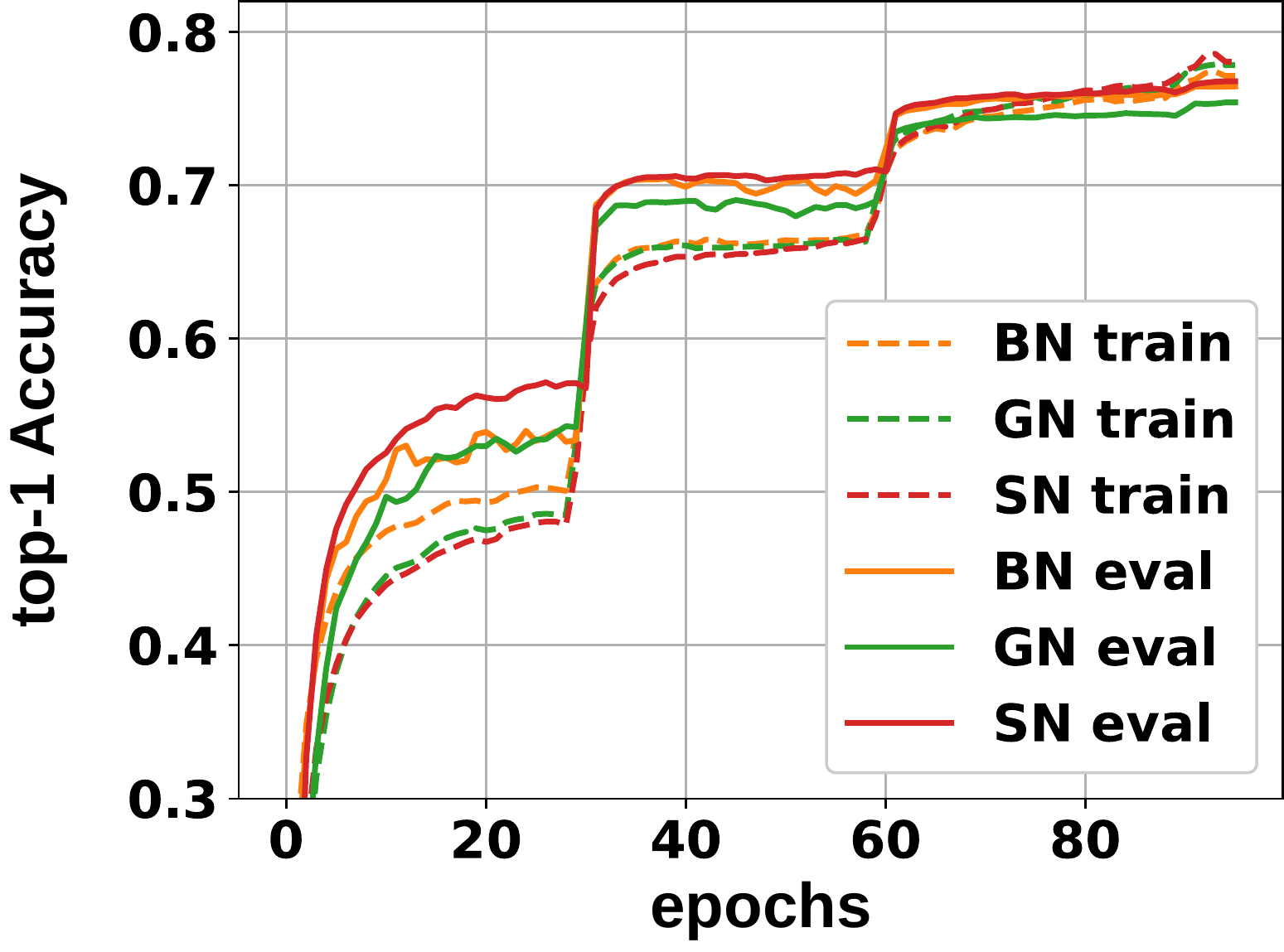}
\end{minipage}%
}%
\hspace{-0.0in}
\subfigure[train and validation curves of (8,2).] {  
\begin{minipage}[c]{0.32\linewidth}
\centering
\includegraphics[height=1.6in]{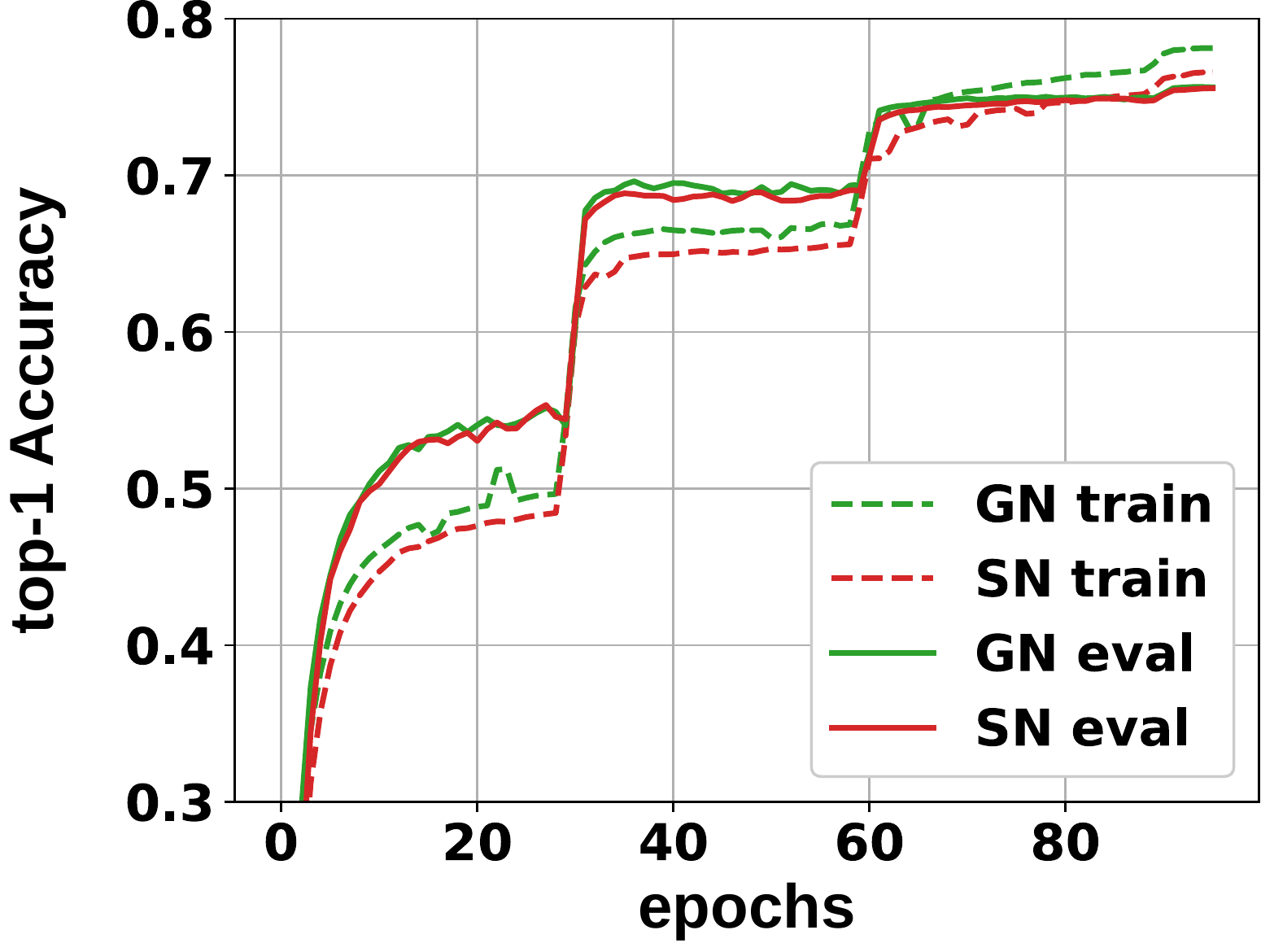}
\end{minipage}
}
\caption{ \small{\textbf{Comparisons of learning curves.} (a) visualizes the validation curves of SN with different settings of batch size. The bracket $(\cdot,\cdot)$ denotes ($\#$GPUs, $\#$samples per GPU). (b) compares the top-1 train and validation curves on ImageNet of SN, BN, and GN in the batch size of (8,32). (c) compares the train and validation curves of SN and GN in the batch size of (8,2).} }
\label{fig:BN-GN-SN}
\end{figure*}

\subsection{Image Classification in ImageNet}\label{sec:imagenet}

SN is first compared with existing normalizers on the ImageNet classification dataset of 1k categories.
All models in ImageNet are trained on 1.2M images and evaluated on 50K validation images.
In the following, we first overview the experimental setting in Sec~\ref{sec:ClassificationSetting}.
The performance comparisons with other popular normalization methods are reported in Sec~\ref{sec:comparison}.
More detailed analysis about different factors and learning dynamics of ratios are presented in Sec~\ref{sec:ablation} and Sec~\ref{sec:learningDynamics}.

\subsubsection{Experimental Setting}
\label{sec:ClassificationSetting}

All the models are trained by using SGD with different settings of batch sizes, which are denoted as a 2-tuple, (\emph{number of GPUs}, \emph{number of samples per GPU}).
For each setting, the gradients are aggregated over all GPUs, and the means and variances of the normalization methods are computed in each GPU.
The network parameters are initialized by following \cite{C:resnet}.
For all normalization methods, all $\gamma$'s
are initialized as 1 and all $\beta$'s as 0.
The parameters of SN ($\lambda_k$ and $\lambda_k^\prime$) are initialized as 1.
We use a weight decay of $10^{-4}$ for all parameters including $\gamma$ and $\beta$.
Without special explanation, all the methods adopt ResNet50 as backbone network and are trained for 100 epoches with a initial learning rate of 0.1, which is deceased by 10$\times$ after 30, 60, and 90 epoches.
For different batch sizes, the initial learning rate is linearly scaled according to \cite{ImageNet1hour}.
During training, we employ data augmentation the same as \cite{C:resnet}.
The top-1 classification accuracy on the 224$\times$224 center crop is reported.

\subsubsection{Performance Comparisons}
\label{sec:comparison}

The top-1 accuracy on the 224$\times$224 center crop is reported for all models.
SN is compared to BN and GN as shown in Table \ref{tab:comp_SN_1}.
%
%
In the first five columns, we see that the accuracy of BN reduces by 1.1\% from $(8,16)$ to $(8,8)$ and declines to 65.3\% of $(8,2)$,
implying that BN is unsuitable in small minibatch, where the random noise from the statistics is too heavy.
GN obtains around 75.9\% in all cases, while SN outperforms BN and GN in almost all cases, rendering its robustness to different batch sizes.
Fig.\ref{fig:BN-GN-SN} plots the training and validation curves,
where SN enables faster convergence while maintains higher or comparable accuracies than those of BN and GN.

The middle two columns of Table \ref{tab:comp_SN_1} average the gradients in a single GPU by using only 16 and 32 samples, such that their batch sizes are the same as $(8,2)$ and $(8,4)$.
SN again performs best in these single-GPU settings, while
BN outperforms GN. For example, unlike $(8,4)$ that uses 8 GPUs,
BN achieves 76.5\% in $(1,32)$, which is the best-performing result of BN,
although the batch size to compute the gradients is as small as 32.
From the above results, we see that BN's performance are sensitive to the statistics more than the gradients, while SN are robust to both of them.
The last two columns of Table \ref{tab:comp_SN_1} have the same batch size of 8, where $(1,8)$ has a minibatch size of 8, while $(8,1)$ is an extreme case with a single sample in a minibatch.
For $(1,8)$, SN performs best.
For $(8,1)$, SN consists of IN and LN but no BN, because IN and BN are the same in training when the minibatch size is 1.
In this case, both SN and GN still perform reasonably well, while BN fails to converge.

Fig.\ref{fig:BN-GN-SN} (a) plots the validation curves of SN. Fig.\ref{fig:BN-GN-SN} (b) and (c) compare the training and validation curves of SN, BN and GN in $(8,32)$ and $(8,2)$ respectively.
From all these curves, we see that SN enables faster convergence while maintains higher or comparable accuracies than those of BN and GN.
On the other hand, SN prevents overfitting problem compared with GN when using both regular (\textit{i.e.} 32 images) and small (\textit{i.e.} 2 images) mini-batch. It benefits from the combination of BN to introduce random noise to promote generalization ability.

In Fig.\ref{fig:BN-GN-SN}, it is noteworthy that the evaluation accuracies are usually higher than training accuracies in the early stage.
This phenomenon may be caused by two factors.
(1) The training accuracy of each epoch is calculated by averaging the loss of each step within the corresponding epoch.  In contrast, the evaluation accuracy is directly tested by using the model of the last step of each epoch. Since the values of accuracies are calculated in a slightly different way, in the early phase of the training, evaluation accuracies are usually higher than the training ones.
(2) The data augmentation (e.g. random crop, multi-scale training, etc.) and dropout are adopted in the training phase but are omitted in the evaluation. In the early stage, these may also cause the difference between training and evaluation accuracies.

\subsubsection{Ablation Study}
\label{sec:ablation}

\begin{figure}[t]
\centering
\includegraphics[width=\linewidth]{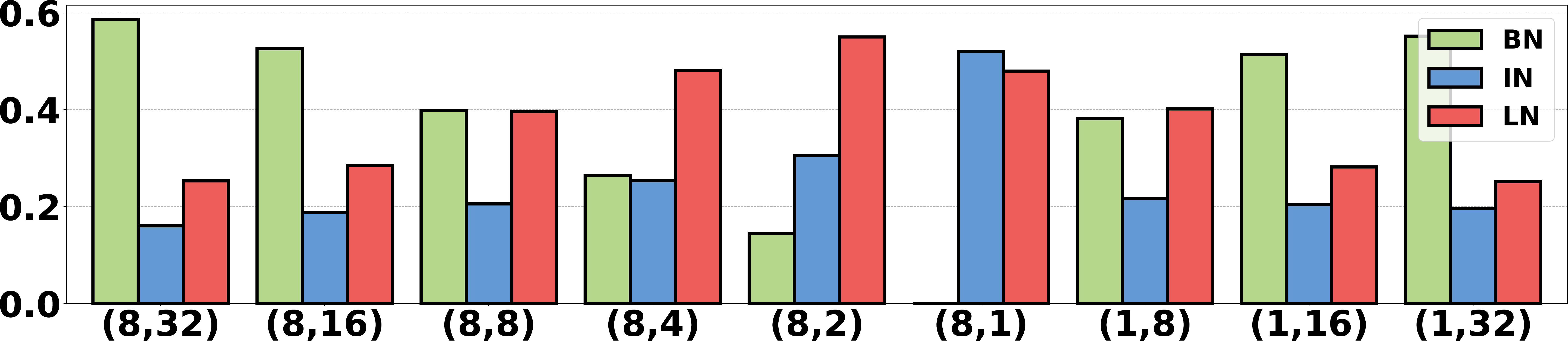}
\caption{\small{Importance weights \vs batch sizes. The bracket $(\cdot,\cdot)$ indicates ($\#$GPUs, $\#$samples per GPU). SN doesn't have BN in $(8,1)$.  \vspace{-10pt}
  }}\label{fig:hist_weight_batch}
\end{figure}

Fig.\ref{fig:intro} (a) and Fig.\ref{fig:hist_weight_batch} plot histograms to compare the importance weights of SN with respect to different tasks and batch sizes.
These histograms are computed by averaging the importance weights of all SN layers in a network.
They show that SN adapts to various scenarios by changing its importance weights.
For example, SN prefers BN when the minibatch is sufficiently large, while it selects LN instead when small minibatch is presented, as shown in the green and red bars of Fig.\ref{fig:hist_weight_batch}. These results are in line with our analyses in Sec.\ref{sec:SN}.

Furthermore, we repeat training of ResNet50 several times in ImageNet, to show that when the network, task, batch setting and data are fixed, the importance weights of SN are not sensitive to the change of training protocols such as solver, parameter initialization, and learning rate decay. As a result, we find that all trained models share similar importance weights.

\begin{figure*}[htbp] \centering
\subfigure[\small{importance weights of $\mu$ in (8,32)}] {
\begin{minipage}[c]{\linewidth}
\centering
\includegraphics[width=0.9\textwidth]{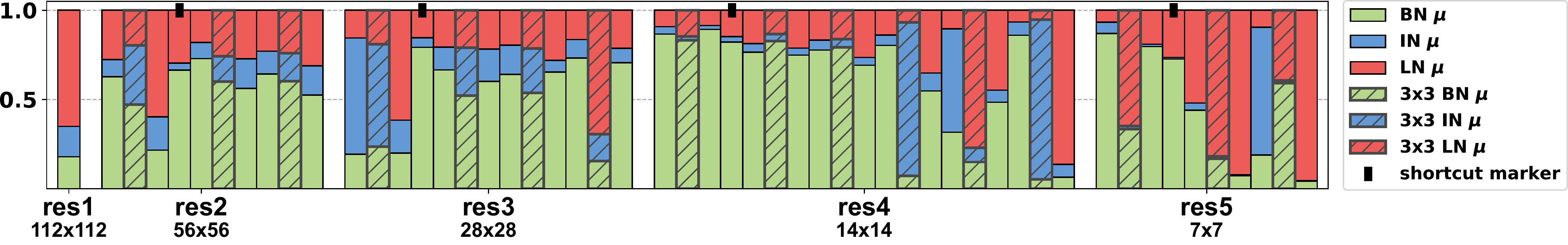}
\end{minipage}%
}%
\hspace{-0.0in}
\subfigure[\small{importance weights of $\sigma$ in (8,32)}] {
\begin{minipage}[c]{\linewidth}
\centering
\vspace{-8pt}
\includegraphics[width=0.9\textwidth]{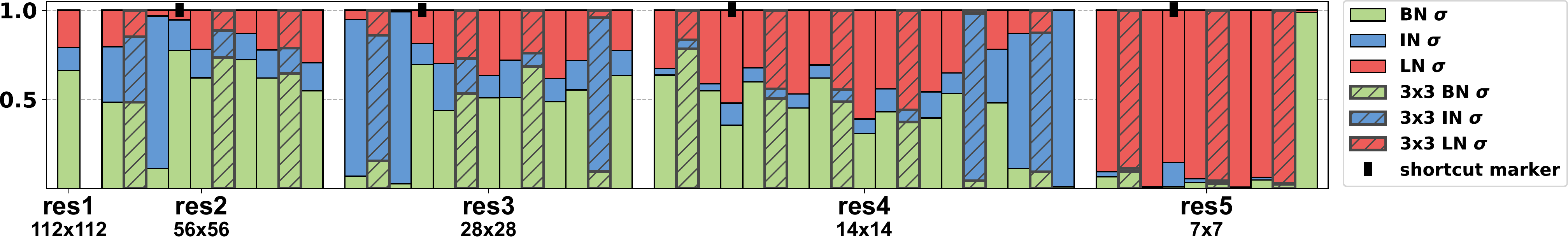}
\end{minipage}%
}%
\hspace{-0.0in}
\subfigure[\small{importance weights of $\mu$ in (8,2)}] {
\begin{minipage}[c]{\linewidth}
\centering
\vspace{-8pt}
\includegraphics[width=0.9\textwidth]{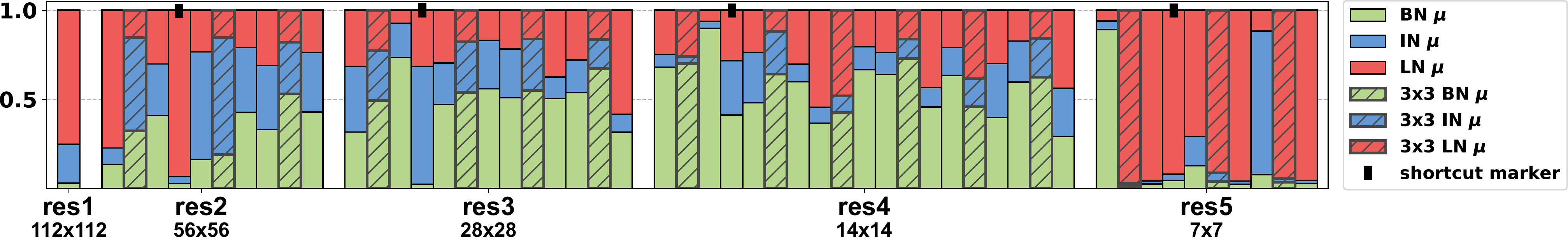}
\end{minipage}%
}%
\hspace{-0.0in}
\subfigure[\small{importance weights of $\sigma$ in (8,2)}] {
\begin{minipage}[c]{\linewidth}
\centering
\vspace{-8pt}
\includegraphics[width=0.9\textwidth]{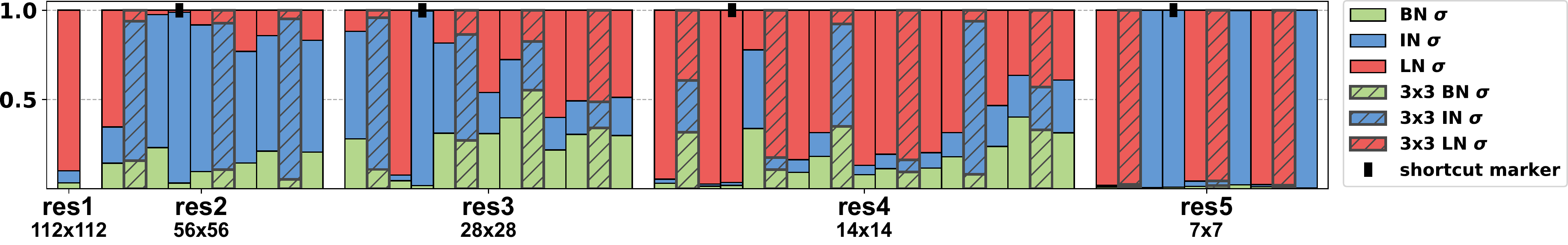}
\end{minipage}%
}%
\vspace{-8pt}
\caption{\small{\textbf{Selected operations of each SN layer in ResNet50.} There are 53 SN layers. (a,b) show the importance weights for $\mu$ and $\sigma$ of $(8,32)$, while (c,d) show those of $(8,2)$.
The $y$-axis represents the importance weights that sum to 1, while the $x$-axis shows different residual blocks of ResNet50.
The SN layers in different places are highlighted differently. For example, the SN layers follow the $3\times3$ conv layers are outlined by shaded color, those in the shortcuts are marked with `$\blacksquare$', while those follow the $1\times1$ conv layers are in flat color.
The first SN layer follows a $7\times7$ conv layer.
We see that SN learns distinct importance weights for different normalization methods as well as $\mu$ and $\sigma$, adapting to different batch sizes, places, and depths of a deep network.}
}
\label{fig:block}
\end{figure*}

The importance weights in each SN layer are visualized in Fig.\ref{fig:block}.
We have several observations to answer what factors that impact the choices of normalizers.
\textbf{First,} for the same batch size, the importance weights of $\mu$ and $\sigma$ could have notable differences, especially when comparing `res1,4,5' of (a,b) and `res2,4,5' of (c,d).
For example, $\sigma$ of BN (green) in `res5' in (b,d) are mostly reduced compared to $\mu$ of BN in (a,c).
As discussed in \cite{C:BN,WN}, this is because the variance estimated in a minibatch produces larger noise than the mean, making training instable.
SN is able to restrain the noisy statistics and stabilize training.
\textbf{Second,} the SN layers in different places of a network may select distinct operations. In other words, when comparing the adjacent SN layers after the $3\times3$ conv layer, shortcut, and the $1\times1$ conv layer, we see that they may choose different importance weights, \eg `res2,3'.
The selectivity of operations in different places (normalization layers) of a deep network has not been observed in previous work.
\textbf{Third,} deeper layers prefer LN and IN more than BN, as illustrated in `res5', which tells us that putting BN in an appropriate place is crucial in the design of network architecture.
Although the stochastic uncertainty in BN (\ie the minibatch statistics) acts as a regularizer that might benefit generalization, using BN uniformly in all normalization layers may impede performance.

\begin{figure}[h]
\centering
\includegraphics[width=0.7\linewidth]{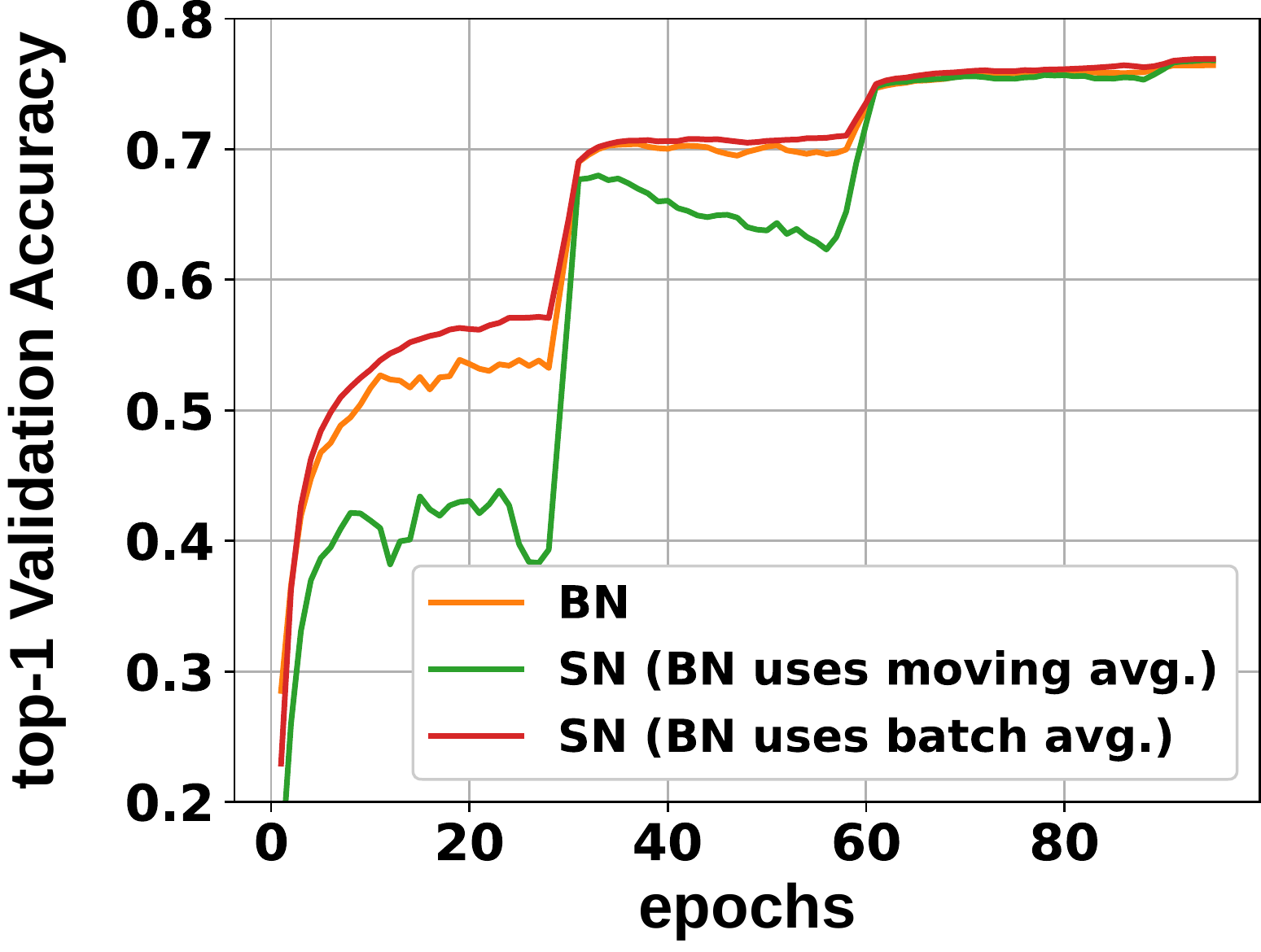}
\caption{\small{Comparisons of `BN', `SN with moving average', and `SN with batch average', when training ResNet50 on ImageNet in $(8,32)$. We see that SN with batch average produces more stable convergence than the other methods.
}}\label{fig:average}
\end{figure}

\textbf{Inference of SN.}
In SN, BN employs batch average rather than moving average. We provide comparisons between them as shown in Fig.\ref{fig:average}, where SN is evaluated with both moving average and batch average to estimate the statistics used in test. They are used to train ResNet50 on ImageNet. The two settings of SN produce similar results of 76.9\% when converged, which is better than 76.4\% of BN.
We see that SN with batch average converges more stably than BN and SN that use moving average.
In this work, we also find that for all batch settings, SN with the batch average provides results better than the moving average.

\textbf{SN \vs IN and LN.}
IN and LN are not optimal in image classification as reported in \cite{IN} and \cite{LN}.
With a regular setting of $(8,32)$, ResNet50 trained with IN and LN achieve 71.6\% and 74.7\% respectively, which reduce 5.3\% and 2.2\% compared to 76.9\% of SN.

\textbf{SN \vs BRN and BKN.}
BRN has two extra hyper-parameters, $r_{max}$ and $d_{max}$, which renormalize the means and variances.
We choose their values as $r_{max}=1.5$ and $d_{max}=0.5$, which work best for ResNet50
in the setting of $(8,4)$ following \cite{BRN}. 73.7\% of BRN surpasses 72.7\% of BN by 1\%, but it reduces 2.2\% compared to 75.9\% of SN.%

BKN \cite{BKN} estimated the statistics in the current layer by combining those computed in the preceding layers. It estimates the covariance matrix rather than the variance vector. In particular, how to connect the layers requires careful design for every specific network.
For ResNet50 with $(8,32)$, BKN achieved 76.8\%, which is comparable to 76.9\% of SN. However, for small minibatch, BKN reported 76.1\% that was evaluated in a micro-batch setting where 256 samples are used to compute gradients and 4 samples to estimate the statistics.
This setting is easier than $(8,4)$ that uses 32 samples to compute gradients.
Furthermore, it is unclear how to apply BRN and BKN in the other tasks such as object detection and segmentation.

\begin{table}[t]
\centering
\begin{tabular}
{c|cc}
\hline
$(8,32)$ &ResNet50 &ResNet101 \\
\hline
BN & 76.4~/~93.0 & 77.6~/~93.6 \\
SN & 76.9~/~93.2& 78.6~/~94.1 \\
SN-BN & 0.5~/~0.2 & 1.0~/~0.5 \\
\hline
\hline
epoch& scratch & finetune\\
\hline
30$^{\mathrm{th}}$ & 76.5~/~93.0 & 77.4~/~93.4 \\
60$^{\mathrm{th}}$ & 76.1~/~93.0 & 77.1~/~93.4 \\
90$^{\mathrm{th}}$ & 76.2~/~93.0 & 76.9~/~93.3\\
\hline
\end{tabular}
\caption{\small{\textbf{Top:} comparisons of top1 / top5 accuracy of BN and SN trained with $(8,32)$ in ImageNet. ``SN-BN'' is the difference between their results. \textbf{Bottom:} comparisons of top1/top5 accuracy between `training from scratch' and `finetuning' ResNet50+SN$(8,32)$ with hard ratios until 100 epochs.
}}\label{tab:comp2}
\end{table}

\subsubsection{Learning Dynamics of Ratios}
\label{sec:learningDynamics}

\begin{figure*}
  \centering
  \includegraphics[width=1.0\linewidth]{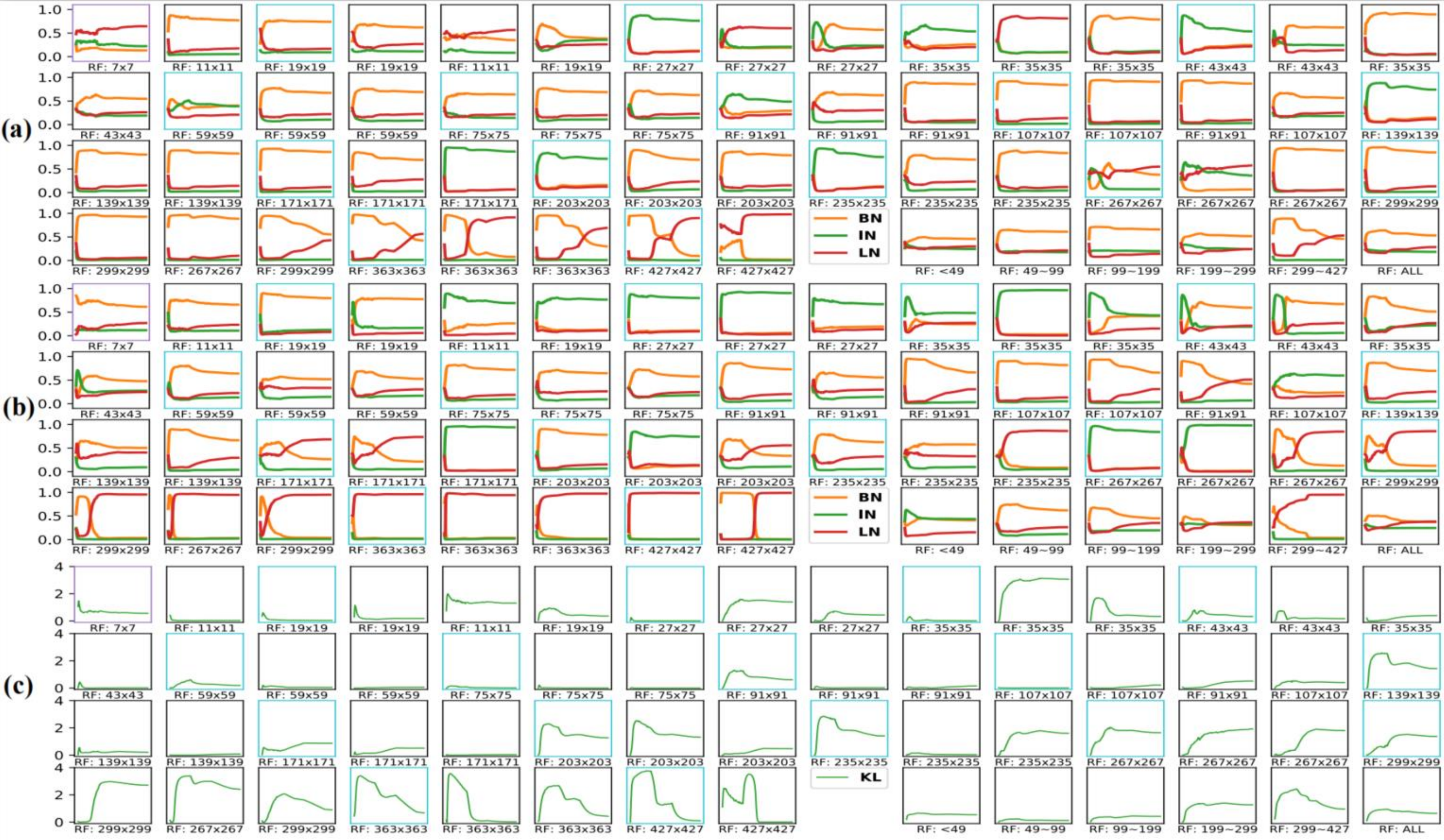}\\
  \caption{\small{Ratios of (a) $\lambda_z^\mu$ and (b) $\lambda_z^\sigma$ in {ResNet50+SN(8,32)} for each normalization layer for 100 epochs, as well as (c) their divergence $\mathcal{D}(\lambda_z^\mu\|\lambda_z^\sigma)$.
  Receptive field (RF) of each layer is given (53 normalization layers in total).
  The last 6 subfigures at the 4$^{\mathrm{th}}$, 8$^{\mathrm{th}}$, and 12$^{\mathrm{th}}$ row show results of different ranges of RF including `RF$<$49', `49$\sim$99', `99$\sim$199', `199$\sim$299', `299$\sim$427', and `ALL' (\ie 7$\sim$427).
 }}\label{fig:SN8-32-mu-sig}
\end{figure*}

\textbf{Soft ratios vary in training.}
The values of soft ratios $\lambda_z^\mu$ and $\lambda_z^\sigma$ are between $0$ and $1$.
%
Fig.\ref{fig:SN8-32-mu-sig}(a,b) plot their values for each normalization layer at every epoch.
These values would have smooth fluctuation in training, implying that different epochs may have their own preference of normalizers.
In general, we see that $\lambda_z^\mu$ mostly prefers BN, while $\lambda_z^\sigma$ prefers IN and BN when receptive field (RF) $<$49 and prefers LN when RF$>$299.
Fig.\ref{fig:SN8-32-mu-sig}(c) shows the discrepancy between (a) and (b) by computing a symmetry metric, that is, $\mathcal{D}(\lambda_z^\mu\|\lambda_z^\sigma)=\mathcal{KL}(\lambda_z^\mu\|\lambda_z^\sigma)
+\mathcal{KL}(\lambda_z^\sigma\|\lambda_z^\mu)$ where $\mathcal{KL}(\cdot\|\cdot)$ is Kullback-Leibler divergence.
Larger value of $\mathcal{D}(\lambda_z^\mu\|\lambda_z^\sigma)$ indicates larger discrepancy between the distributions of $\lambda_z^\mu$ and $\lambda_z^\sigma$.

For example, $\lambda_z^\mu$ and $\lambda_z^\sigma$ of the first layer choose different normalizers (see the first subfigure in (a,b)), making them had moderately large divergence $\mathcal{D}(\lambda_z^\mu\|\lambda_z^\sigma)\approx1$ (see the first subfigure in (c)).
Moreover, the ratios of the 2$^{\mathrm{nd}}$, 3$^{\mathrm{rd}}$, and 4$^{\mathrm{th}}$ layer are similar and they have small divergence $\mathcal{D}(\lambda_z^\mu\|\lambda_z^\sigma)<0.5$.
We also see that $\lambda_z^\mu$ and $\lambda_z^\sigma$ prefer different normalizers when RF is 199$\sim$299 and 299$\sim$427 where $\mathcal{D}(\lambda_z^\mu\|\lambda_z^\sigma)>2$, as shown in
the last row of (c).
In conclusion, more than 50\% number of layers have large divergence between $\lambda_z^\mu$ and $\lambda_z^\sigma$, confirming that different ratios should be learned for $\mu$ and $\sigma$.

\begin{figure}
\centering
\includegraphics[width=1.0\linewidth]{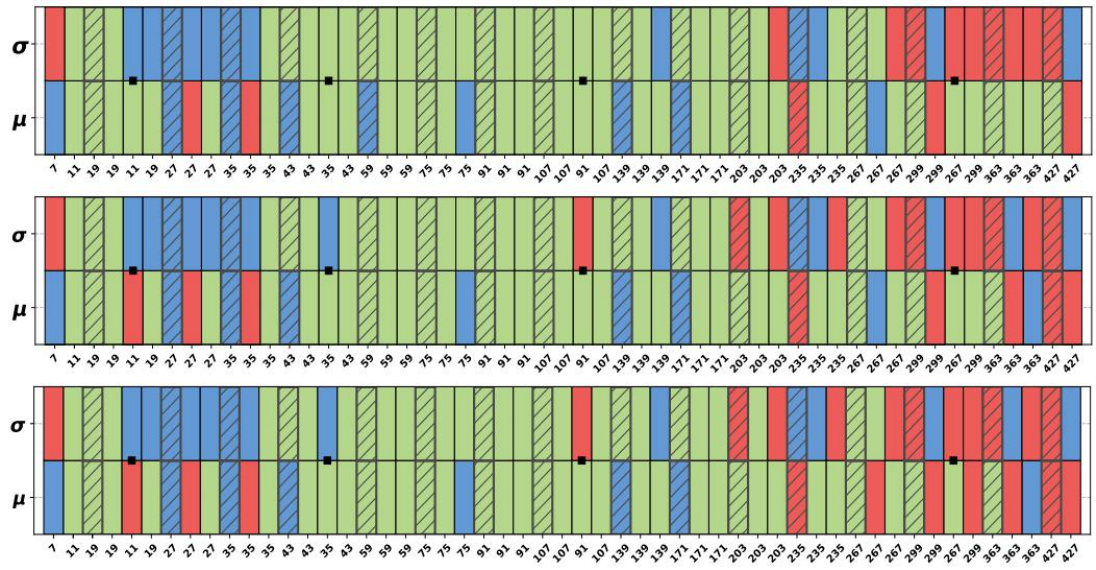}\\
\caption{{\small \textbf{Hard ratios} for variance ($\sigma$) and mean ($\mu$) including BN ({\color{green}\textbf{green}}), IN ({\color{blue}\textbf{blue}}), and LN ({\color{red}\textbf{red}}). Snapshots of ResNet50 trained after 30 (\textbf{top}), 60 (\textbf{middle}), and 90 (\textbf{bottom}) epochs are shown.
RF is given for each layer (53 normalization layers in total). A bar with slashes denotes SN after $3\times3$ conv layer (the others are $1\times1$ conv). A black square `$\blacksquare$' indicates SN at the shortcut. It's better to zoom in 200\%.
}}\label{fig:SN8-32-epoch}
\end{figure}

\textbf{Hard ratios are relatively stable,} although the soft ratios are varying in training.
A hard ratio is a sparse vector obtained by applying $\mathrm{max}$ function to $\lambda_z^\mu$ or $\lambda_z^\sigma$ such as $\mathrm{max}(\lambda_z^\sigma)$, that is, only one entry is `1' and the others are `0' to select only one normalizer.
Fig.\ref{fig:SN8-32-epoch} shows hard ratios for each layer in three snapshots, which are ResNet50+SN(8,32) trained after 30, 60, and 90 epochs respectively.
For example, $\sigma$ and $\mu$ use LN and IN respectively in the first layer in Fig.\ref{fig:SN8-32-epoch}.

We have several observations. First, the size of filters ($3\times3$ or $1\times1$) seem to have no preference of specific normalizer, while the skip-connections prefer different normalizers for $\sigma$ and $\mu$ at the 90$^{\mathrm{th}}$ epoch.
Second, around $50\%$ number of layers select two different normalizers for $\sigma$ and $\mu$ in these three snapshots, which are $49\%$ ($26/53$), $53\%$ ($28/53$), and $51\%$ ($27/53$) respectively.
Third, the discrepancy between snapshots are small. For instance, $10$ layers are different when comparing between 30$^{\mathrm{th}}$ and 60$^{\mathrm{th}}$ epoch, while only $2$ layers are different between 60$^{\mathrm{th}}$ and 90$^{\mathrm{th}}$ epoch.
Fourth, the layers that choose different normalizers are mainly presented when RF$<$40 and $>$200, rendering depth would be a major factor that affects the ratios.

\textbf{Performance of hard ratios.}
We further examine performance of hard ratios.
%
We finetune the above snapshots by replacing soft ratios with hard ratios.
The right of Table \ref{tab:comp2} shows that these models achieve slightly better performance compared with their soft counterpart ($76.9/93.3$).
An intuitive explanation is that sparseness can effectively prevent the model from overfitting.
The similar results are also presented in the recent proposed Sparse Switchable Normalization (SSN)~\cite{C:ssn}.
It implies that we could increase sparsity in ratios to reduce computations of multiple normalizers while maintaining good performance.
%

Furthermore, we train the above models from scratch by initializing the ratios as the hard ratios in the above snapshots, rather than using the default initial value $\frac{1}{3}$.
However, this setting harms generalization as shown in Table \ref{tab:comp2}. We conjecture that all normalizers in $\Omega$ are helpful to smooth loss landscape at the beginning of training. Therefore, the sparsity of ratios should be enhanced gently as training progresses.
In other words, initializing ratios by $\frac{1}{3}$ is a good practice. Tuning this value is cumbersome and may also imped generalization ability.


\subsection{Object Detection and Instance Segmentation}\label{sec:coco}

Next we evaluate SN in
object detection and instance segmentation in COCO \cite{lin2014microsoft}.
Unlike image classification, these two tasks benefit from large size of input images, making large memory footprint and therefore leading to small minibatch size, such as 2 samples per GPU \cite{FasterRCNN,FPN}.
In this case, as BN is not applicable in small minibatch,
previous work \cite{FasterRCNN,FPN,MaskRCNN} often
freeze BN and turns it into a constant linear transformation layer, which actually performs no normalization.
Overall, SN selects different operations in different components of a detection system (see Fig.\ref{fig:intro}), showing much more superiority than both BN and GN.

\subsubsection{Experimental Settings}
\label{sec:DetectionSettings}

In practice, we implement object detection and instance segmentation on existing detection softwares of PyTorch and Caffe2-Detectron \cite{detectron} respectively.
We conduct 3 settings, including \textbf{setting-1:} Faster R-CNN \cite{FasterRCNN} on PyTorch; \textbf{setting-2:} Faster R-CNN+FPN \cite{FPN} on Caffe2; and \textbf{setting-3:} Mask R-CNN \cite{MaskRCNN}+FPN on Caffe2.
For all these settings, we choose ResNet50 as the backbone network.
In each setting,
the experimental configurations of all the models are the same, while only the normalization layers are replaced.
All models of SN are finetuned from $(8,2)$ in ImageNet.
For \textbf{setting-1}, we employ a fast implementation \cite{jjfaster2rcnn} of Faster R-CNN in PyTorch and follow its protocol. Specifically,
we train all models on 4 GPUs and 3 images per GPU. Each image is re-scaled such that its shorter side is 600 pixels.
%
All models are trained for 80k iterations with a learning rate of 0.01 and then for another 40k iterations with 0.001.
For \textbf{setting-2} and \textbf{setting-3}, we employ the configurations of the Caffe2-Detectron \cite{detectron}. We train all models on 8 GPUs and 2 images per GPU. Each image is re-scaled to its shorter side of 800 pixels.
%
In particular, for setting-2, the learning rate (LR) is initialized as 0.02 and is decreased by a factor of 0.1 after 60k and 80k iterations and finally terminates at 90k iterations. This is referred as the 1$\mathrm{x}$ schedule in Detectron.
In setting-3, the LR schedule is twice as long as the 1$\mathrm{x}$ schedule with the LR decay points scaled twofold proportionally, referred as 2$\mathrm{x}$ schedule.
For all settings, we set weight decay to 0 for both $\gamma$ and $\beta$ following \cite{GN}.

All of the models
are trained in the \emph{2017 train} set of COCO by using SGD with a momentum of 0.9 and a weight decay of $10^{-4}$ on the network parameters, and tested in the \emph{2017 val} set.
We report the standard metrics of COCO, including average precisions at IoU=0.5:0.05:0.75 (AP), IoU=0.5 (AP$_{.5}$), and IoU=0.75 (AP$_{.75}$) for both bounding box (AP$^{\mathrm{b}}$) and segmentation mask (AP$^{\mathrm{m}}$). Also, we report average precisions for small (AP$_s$), medium (AP$_m$), and large (AP$_l$) objects.

\begin{table}
\centering
\begin{tabular}{ll|ccc|ccc}
  \hline
 \tabincell{c}{backbone} & head & AP & AP$_{{.5}}$ & AP$_{{.75}}$ & AP$_{{l}}$ & AP$_{{m}}$ & AP$_{{s}}$ \\
  \hline
    BN$^\dag$ & BN$^\dag$ & 29.6 & 47.8 & 31.9 & 45.5 & 33.0 &11.5 \\
    BN & BN & 19.3 & 33.0 & 20.0 & 32.3 & 21.3 & 7.4\\
    GN & GN & 32.7 & 52.4 & 35.1 & \textbf{49.1} & 36.1 & 14.9 \\
    SN & SN & \textbf{33.0} & \textbf{52.9} & \textbf{35.7} & 48.7 & \textbf{37.2} & \textbf{15.6}\\
    \hline
    BN$^\ddagger$ & BN & 20.0 & 33.5 & 21.1 & 32.1 & 21.9 & 7.3\\
    GN$^\ddagger$ & GN & 28.3 & 46.3 & 30.1 & 41.2 & 30.0 &  12.7 \\
    SN$^\ddagger$ & SN & \textbf{29.5} & \textbf{47.8} & \textbf{31.6} & \textbf{44.2} & \textbf{32.6} & \textbf{13.0}\\
  \hline
  \vspace{-10pt}
  \end{tabular}
  \caption{\small{\textbf{Faster R-CNN for detection in COCO} using
ResNet50 and RPN. BN$^\dag$ represents BN is frozen without finetuning.
The superscript `$\ddagger$' indicates the backbones are trained from scratch without pretraining on ImageNet.
The best results in the upper and lower parts are bold.
}}\label{tab:det_faster}
\end{table}

\subsubsection{Performance Comparisons}
\label{sec:DetectionComparison}

\textbf{Results of Faster R-CNN.} As shown in Table \ref{tab:det_faster}, SN is compared with both BN and GN in the Faster R-CNN.
In this setting, the layers up to conv4 of ResNet50 are used as \emph{backbone} to extract features, and the layers of conv5 are used as the Region-of-Interest \emph{head} for classification and regression.
As the layers are inherited from the pretrained model, both the backbone and head involve normalization layers.
Different results of Table \ref{tab:det_faster} use different normalization methods in the backbone and head.
Its upper part shows results of finetuning the ResNet50 models pretrained
on ImageNet. The lower part compares training COCO from scratch without pretraining on ImageNet.

\begin{figure}[t] \centering
\subfigure[Finetuning.] {
\begin{minipage}[c]{0.48\linewidth}
\centering
\includegraphics[width=1\textwidth]{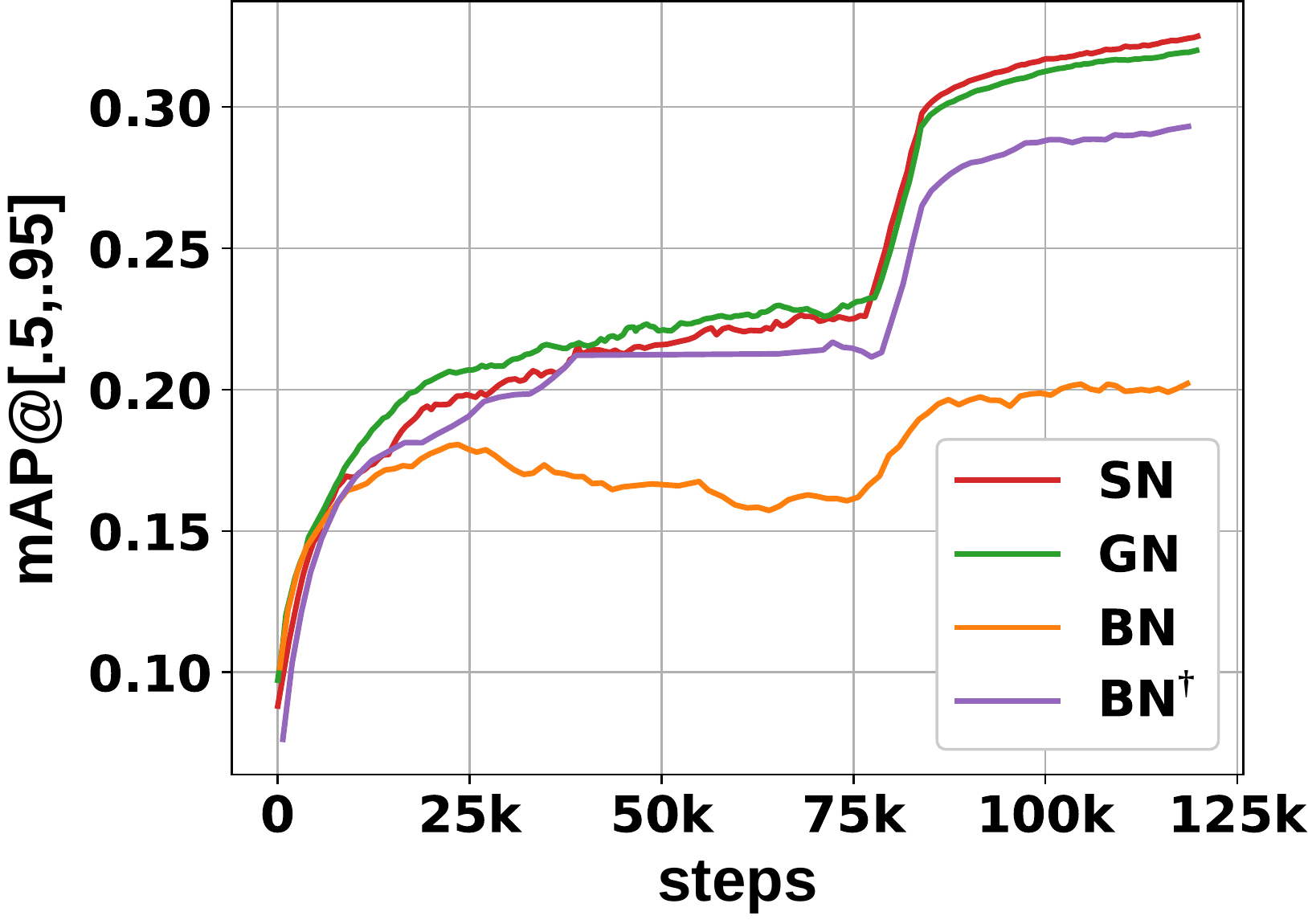}
\end{minipage}%
}%
\hspace{-0.0in}
\subfigure[Training from Scratch.] {
\begin{minipage}[c]{0.48\linewidth}
\centering
\includegraphics[width=1\textwidth]{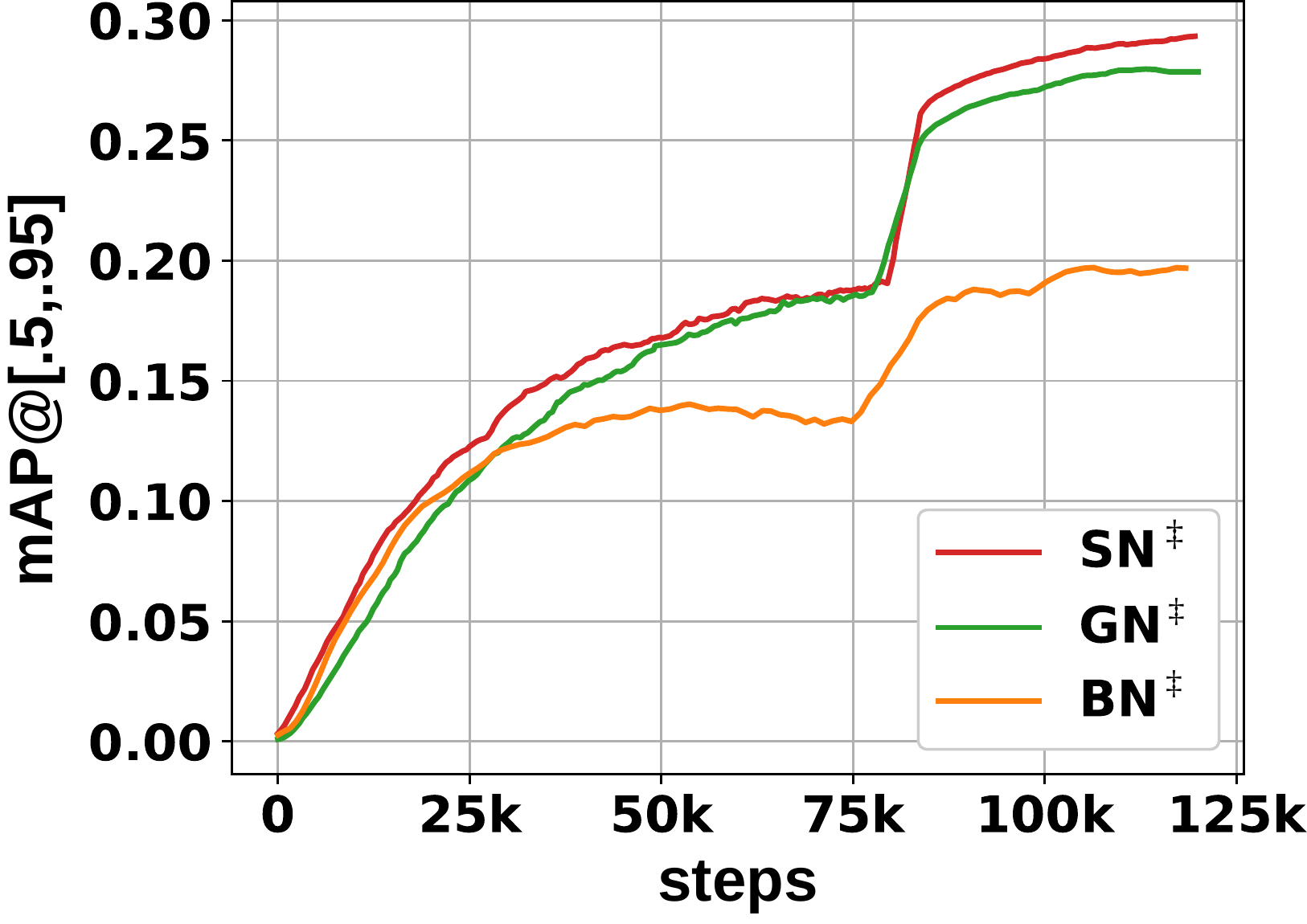}
\end{minipage}%
}%
\caption{ \small{Average precision (AP) curves of Faster R-CNN on the \emph{2017 val} set of COCO. (a) plots the results of finetuning pretrained networks. (b) shows training the models from scratch.}
\hspace{-15pt}
}
\label{fig:faster-rcnn}
\end{figure}

In the upper part of Table \ref{tab:det_faster}, the baseline is denoted as BN$^\dag$, where the BN layers are frozen.
We see that freezing BN performs significantly better than finetuning BN (29.6 \vs 19.3).
SN and GN enable finetuning the normalization layers, where SN obtains the best-performing AP of 33.0 in this setting.
Fig.\ref{fig:faster-rcnn} (a) compares their AP curves.
As reported in the lower part of Table \ref{tab:det_faster}, SN and GN allow us to train COCO from scratch without pretraining on ImageNet, and they still achieve competitive results.
For instance, 29.5 of SN$^\ddagger$ outperforms BN$^\ddagger$ by a large margin of 9.5 AP and GN$^\ddagger$ by 1.2 AP.
Their learning curves are compared in Fig.\ref{fig:faster-rcnn} (b).

\textbf{Results of Faster R-CNN + FPN.}
Table \ref{tab:det_faster_FCN} reports results of Faster R-CNN by using ResNet50 and the Feature Pyramid Network (FPN) \cite{FPN}.
%
%
%
%
A baseline BN$^\dagger$ achieves an AP of 36.7 without using normalization in the detection head.
When using SN and GN in the head and BN$^\dagger$ in the backbone, BN$^\dagger$+SN improves the AP of BN$^\dagger$+GN by 0.8 (from 37.2 to 38.0).
We investigate using SN and GN in both the backbone and head. In this case, we find that GN improves BN$^\dagger$+SN by only a small margin of 0.2 AP (38.2 \vs 38.0), although the backbone is pretrained and finetuned by using GN.
When finetuning the SN backbone, SN obtains a significant improvement of 1.1 AP over GN (39.3 \vs 38.2).
Furthermore, the 39.3 AP of SN and 38.2 of GN both outperform 37.8 in \cite{MegDet}, which synchronizes BN layers in the backbone (\ie BN layers are not frozen).

\begin{table}[t]
\centering
\begin{tabular}{p{25pt}<{\centering}p{15pt}<{\centering}|p{10pt}<{\centering}p{10pt}<{\centering}
p{18pt}<{\centering}|p{10pt}<{\centering}p{10pt}<{\centering}p{18pt}<{\centering}}
\hline
\tabincell{c}{backbone} & head & AP & AP$_{.5}$ & AP$_{.75}$ & AP$_{{l}}$ & AP$_{{m}}$ & AP$_{{s}}$ \\
\hline
BN$^\dag$ & -- & 36.7 & 58.4  & 39.6 & 48.1 & 39.8 & 21.1\\
BN$^\dag$ & GN & 37.2 & 58.0 & 40.4 & 48.6 & 40.3 & 21.6\\
BN$^\dag$ & SN & 38.0 & 59.4 & 41.5 & 48.9 &  41.3& {22.7}\\
\hline
GN & GN & 38.2 & 58.7 & 41.3 & 49.6 & 41.0 &  22.4\\
SN & SN & \textbf{39.3} & \textbf{60.9} & \textbf{42.8} & \textbf{50.3} & \textbf{42.7} &\textbf{23.5}\\
\hline
\end{tabular}
\caption{\small{\textbf{Faster R-CNN+FPN} using
ResNet50 and FPN with 1$\mathrm{x}$ LR schedule. BN$^\dag$ represents BN is frozen.
The best results are bold.
}}\label{tab:det_faster_FCN}
\end{table}

\begin{figure*}[t]
\centering
\includegraphics[width=\textwidth,height=100mm]{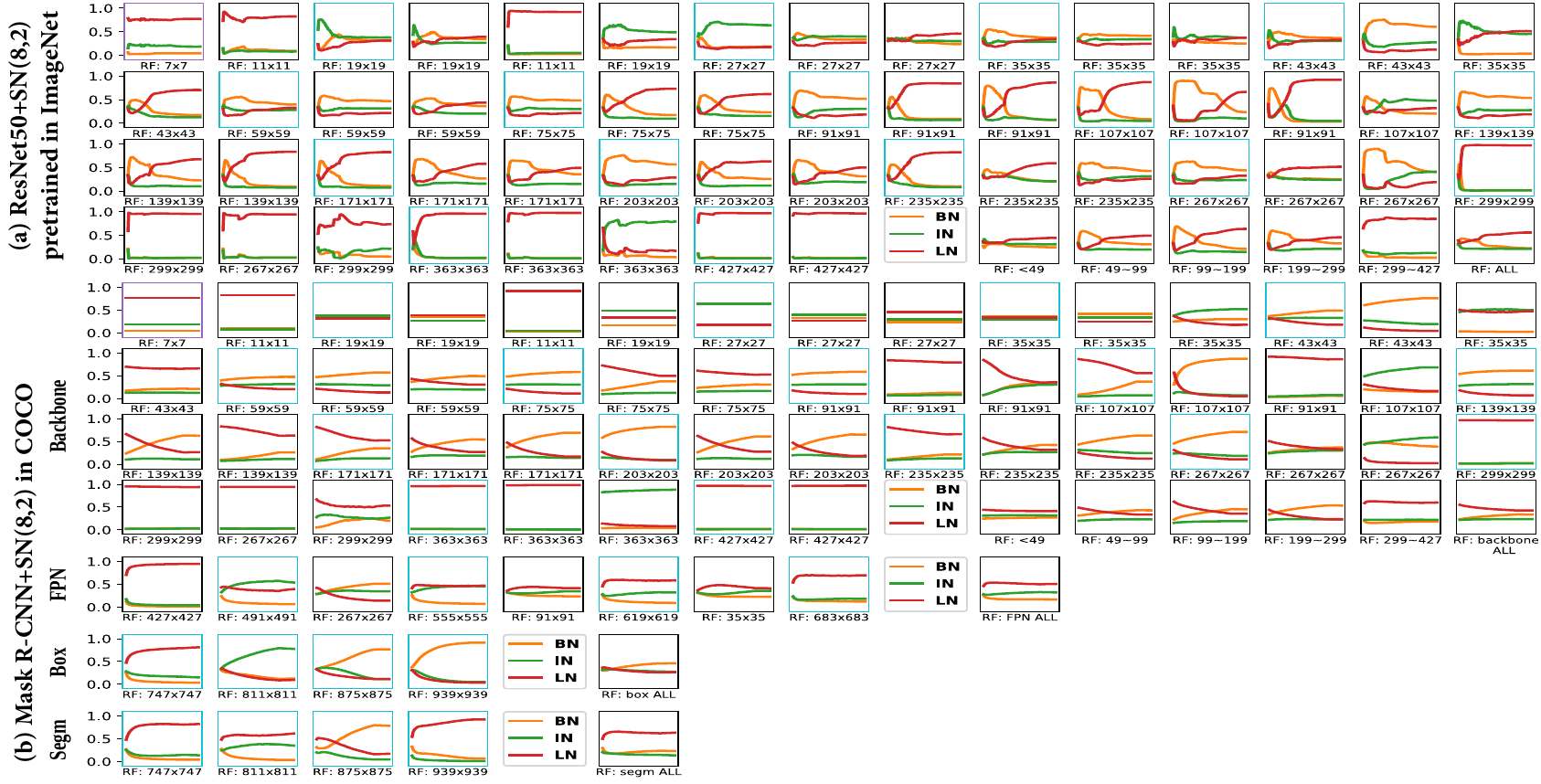}
\vspace{-10pt}
\caption{\small{(a) shows ratios of ResNet50+SN$(8,2)$ in ImageNet. (b) shows ratios of Mask R-CNN+SN$(8,2)$ when finetuning in COCO including backbone (ResNet50), FPN, box stream, and mask stream. \vspace{-8pt}
}}
\label{fig:det_coco}
\end{figure*}

\begin{figure*}[htbp]\centering
\subfigure{
\begin{minipage}[c]{1.0\linewidth}
\centering
\includegraphics[width=1\textwidth,height=24mm]{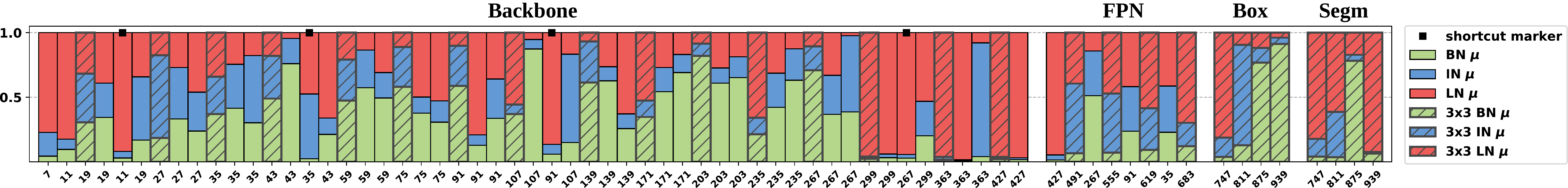} \\
\includegraphics[width=1\textwidth,height=24mm]{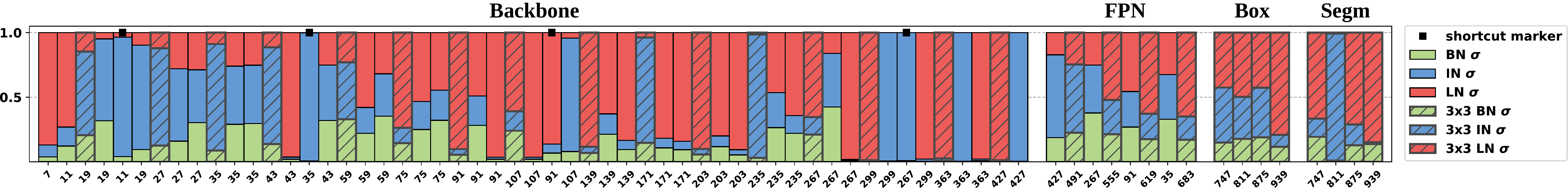}
\end{minipage}%
}%
\vspace{-10pt}
\caption{{$\lambda_z^\mu$ (top) and $\lambda_z^\sigma$ (bottom) of Mask R-CNN+SN(8,2) are shown when training converged in COCO, including backbone (ResNet50), FPN, box stream, and mask stream.
}}
\label{fig:det_converge}
\end{figure*}

\begin{table}[t]
\centering
\begin{tabular}{p{25pt}<{\centering}p{15pt}<{\centering}|p{10pt}<{\centering}p{10pt}<{\centering}
p{18pt}<{\centering}|p{10pt}<{\centering}p{10pt}<{\centering}p{10pt}<{\centering}}
  \hline
 \tabincell{c}{backbone} & head & AP$^{\mathrm{b}}$ & AP$^{\mathrm{b}}_{.5}$ & AP$^{\mathrm{b}}_{.75}$ & AP$^{\mathrm{m}}$ & AP$^{\mathrm{m}}_{.5}$ & AP$^{\mathrm{m}}_{.75}$ \\
  \hline
BN$^\dag$ & -- & 38.6 & 59.5 & 41.9 & 34.2 & 56.2 & 36.1\\
 BN$^\dag$ & GN & 39.5 & 60.0 & 43.2 & 34.4 & 56.4 & 36.3\\
 BN$^\dag$ & SN & 40.0 & 61.0 & 43.3 & 34.8 & 57.3 & 36.3 \\
 \hline
 GN & GN & 40.2 & 60.9& 43.8 & 35.7 & 57.8 & 38.0 \\
 GN & SN & {40.4} & {61.4} & {44.2} &{36.0} & {58.4} & {38.1}\\
 SN & SN & \textbf{41.0} & \textbf{62.3} & \textbf{45.1} &\textbf{36.5} & \textbf{58.9} & \textbf{38.7}\\
  \hline
\end{tabular}
\caption{\small{\textbf{Mask R-CNN} using
ResNet50 and FPN with 2$\mathrm{x}$ LR schedule. BN$^\dag$ represents BN is frozen without finetuning. The best results are bold.
}}\label{tab:det_mask}
\end{table}

\textbf{Results of Mask R-CNN + FPN.}
Table \ref{tab:det_mask} reports results of Mask R-CNN \cite{MaskRCNN} with FPN. In the upper part, SN is compared to a head with no normalization and a head with GN, while the backbone is pretrained with BN, which is then frozen in finetuning (\ie the ImageNet pretrained features are the same).
We see that the baseline BN$^\dagger$ achieves a box AP of 38.6 and a mask AP of 34.2.
SN improves GN by 0.5 box AP and 0.4 mask AP, when finetuning the same BN$^\dagger$ backbone.

More direct comparisons with GN are shown in the lower part of Table \ref{tab:det_mask}. We apply SN in the head and finetune the same backbone network pretrained with GN.
In this case, SN outperforms GN by 0.2 and 0.3 box and mask APs respectively.
Moreover, when finetuning the SN backbone, SN surpasses GN by a large margin of both box and mask AP (41.0 \vs 40.2 and 36.5 \vs 35.7).
Note that the performance of SN even outperforms 40.9 and 36.4 of the 101-layered ResNet \cite{detectron}.



\textbf{Analysis of Dynamics of Ratios.}
For object detection in COCO, we find that the ratio of BN gradually increases when 49$<$RF$<$299 ($>$0.5). This could be attributed to the two-stage pipeline of Mask R-CNN.
To see this, Fig.\ref{fig:det_coco}(b) and Fig.\ref{fig:det_converge} plot the ratios in COCO.
We see that BN has larger impact in backbone and box stream than in FPN and mask stream.
It demonstrates the regularization effect of BN could change depending on the tasks (\textit{e.g.} bounding-box based or mask based) and network architectures design.
%
Furthermore, ratios in the backbone have different dynamics in pretraining and finetuning by comparing two ResNet50 models in Fig.\ref{fig:det_coco}(a,b), that is, the BN ratios decrease in pretraining for recognition, while increase in finetuning for detection even though the batch size is $(8,2)$.

\subsection{Semantic Image Segmentation}\label{sec:segmentation}

We investigate SN in semantic image segmentation in ADE20K~\cite{C:ADE20K} and Cityscapes~\cite{C:cityscape}.
Similar to object detection, semantic image segmentation also benefits from large input size, making the minibatch size is small during training. We use 2 samples per GPU for both of the datasets.
We employ the open-source software in PyTorch\footnote{\url{https://github.com/CSAILVision/semantic-segmentation-pytorch}} and only replace the normalization layers in CNNs with the other settings fixed.

\subsubsection{Experimental Settings}
\label{sec:SegSettings}

\begin{figure*}[htbp] \centering
\subfigure[\small{importance weights of $\mu$ of (8,2) in ADE20K.}] {
\begin{minipage}[c]{\linewidth}
\centering
\includegraphics[width=0.9\textwidth]{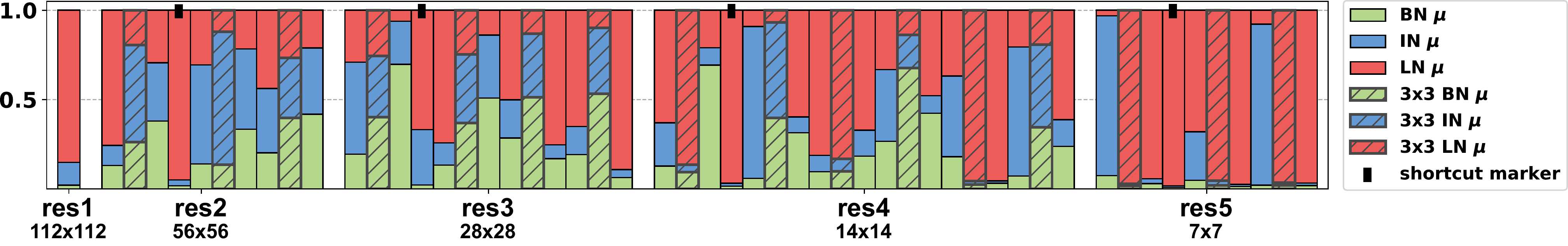}
\end{minipage}%
}%
\hspace{-0.0in}
\subfigure[\small{importance weights of $\sigma$ of (8,2) in ADE20K.}] {
\begin{minipage}[c]{\linewidth}
\centering
\vspace{-8pt}
\includegraphics[width=0.9\textwidth]{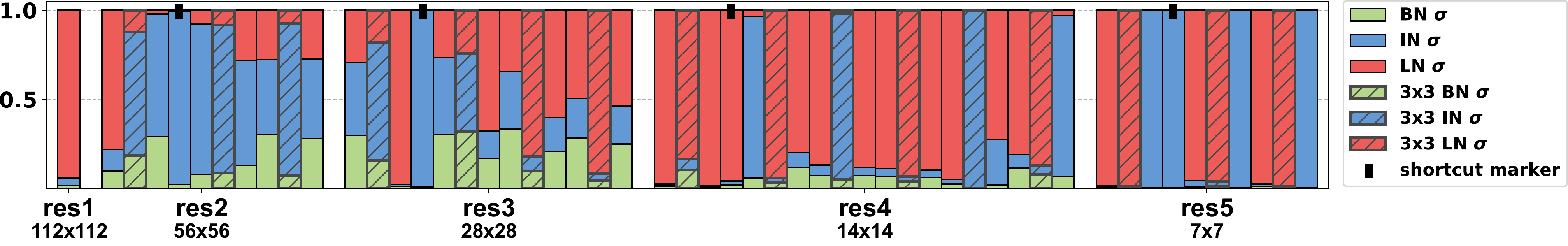}
\end{minipage}%
}%
\hspace{-0.0in}
\subfigure[\small{importance weights of $\mu$ of (8,2) in Cityscapes.}] {
\begin{minipage}[c]{\linewidth}
\centering
\vspace{-8pt}
\includegraphics[width=0.9\textwidth]{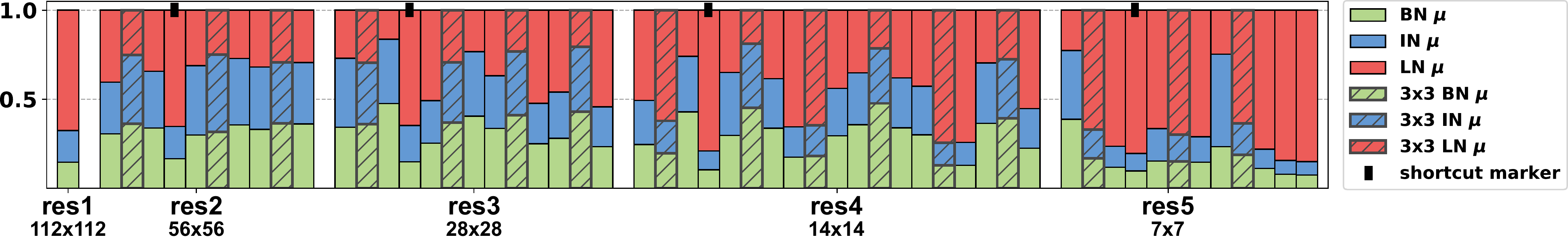}
\end{minipage}%
}%
\hspace{-0.0in}
\subfigure[\small{importance weights of $\sigma$ of (8,2) in Cityscapes.}] {
\begin{minipage}[c]{\linewidth}
\centering
\vspace{-8pt}
\includegraphics[width=0.9\textwidth]{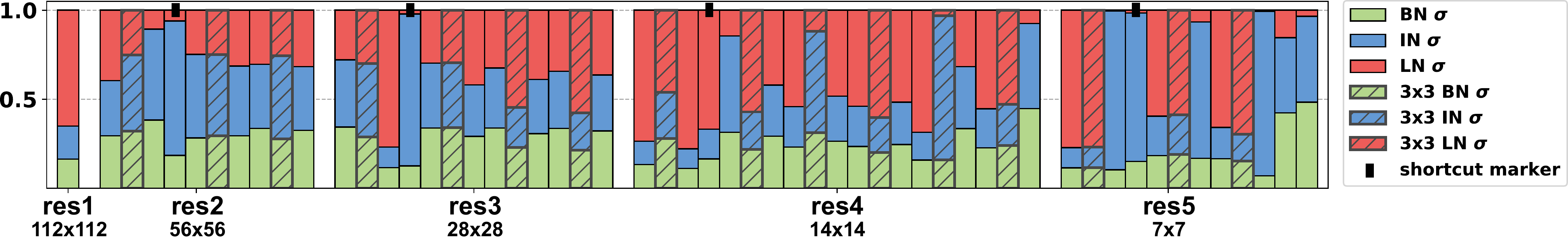}
\end{minipage}%
}%
\caption{\small{\textbf{Selected normalizers of each SN layer in ResNet50 for semantic image parsing in ADE20K and Cityscapes}. There are 53 SN layers. (a,b) show the importance weights for $\mu$ and $\sigma$ of $(8,2)$ in ADE20K, while (c,d) show those of $(8,2)$ in Cityscapes.
The $y$-axis represents the importance weights that sum to 1, while the $x$-axis shows different residual blocks of ResNet50.
The SN layers in different places are highlighted differently. For example, the SN layers follow the $3\times3$ conv layers are outlined by shaded color, those in the shortcuts are marked with `$\blacksquare$', while those follow the $1\times1$ conv layers are in flat color.
%
%
}
\vspace{-5pt}}
\label{fig:segm_block}
\end{figure*}


For both datasets, we use  DeepLab~\cite{J:DeepLab} with ResNet50 as the backbone network, where
%
$\mathrm{output}\_\mathrm{stride}=8$ and the last two blocks in the original ResNet contains atrous convolution with $\mathrm{rate}=2$ and $\mathrm{rate}=4$ respectively.
Following~\cite{C:PSPNet}, we employ ``poly" learning rate policy with $\mathrm{power}=0.9$ and use the auxiliary loss with the weight $0.4$ during training.
%
%
The bilinear operation is adopted to upsmaple the score maps in the validation phase.

\textbf{ADE20K.}
SyncBN and GN adopt the pretrained models on ImageNet.
SyncBN collects the statistics from $8$ GPUs. Thus the actual ``batchsize" is $16$ during training.
To evaluate the performance of SN and SyncSN, we use SN $(8,2)$, $(8,4)$, $(8,32)$ in ImageNet as the pretrained model, respectively.
%
For all models, we resize each image to $450\times450$ and train for $100,000$ iterations. We performance multi-scale testing with $\mathrm{input}\_\mathrm{size}=\{300,400,500,600\}$.

\textbf{Cityscapes.}
For all models, we finetune from their pretrained ResNet50 models.
Same as ADE20K, both SN and SyncSN finetune from $(8,2)$, $(8,4)$, $(8,32)$ repsectively. For all models, the batchsize is $16$ in finetuning.
We use random crop with the size $713\times713$ and train for $400$ epoches. For multi-scale testing, the inference scales are $\{1.0, 1.25, 1.5, 1.75\}$.
%


\begin{table}[t]
\centering
\begin{tabular}{r |p{28pt}<{\centering}p{28pt}<{\centering}| p{28pt}<{\centering}p{28pt}<{\centering}}
\hline
           & \multicolumn{2}{c|}{  ADE20K }            & \multicolumn{2}{c}{  Cityscapes }  \\

& mIoU$_{\mathrm{ss}}$ & mIoU$_{\mathrm{ms}}$  & mIoU$_{\mathrm{ss}}$   & mIoU$_{\mathrm{ms}}$ \\
\hline
SyncBN   & 36.4 & 37.7 & 69.7  &  73.0\\
GN        &       35.7    &  36.3   & 68.4 &73.1\\
SN~(8,2)        & \textbf{38.7} & \textbf{39.2} & 71.6  & 75.4\\
SN~(8,4)        & 38.6 & 39.0 &  72.1 & \textbf{75.8} \\
SN~(8,32)        & 37.7 & 38.4 & \textbf{72.2}  & \textbf{75.8} \\
\hline
SyncSN~(8,2)    &　39.0　& 39.3  & 75.5 & 76.9\\
SyncSN~(8,4)    &　39.8　& 39.9  & 75.8 & 77.0 \\
SyncSN~(8,32)    & \textbf{40.0}　& \textbf{40.4}  & \textbf{76.7} & \textbf{77.5} \\
\hline
\end{tabular}
\caption{\small{\textbf{Results in ADE20K validation set and Cityscapes test set} by using
ResNet50 with dilated convolutions. `$\mathrm{ss}$' and `$\mathrm{ms}$' indicate single-scale and multi-scale inference. SyncBN represents mutli-GPU synchronization of BN. SyncSN indicates the BN in SN is synchronized across mutli-GPU.
}}\label{tab:segmentation}
\end{table}


\begin{table}[t]
\centering
\begin{tabular}{l |p{20pt}<{\centering}p{18pt}<{\centering}|p{11pt}<{\centering}p{11pt}<{\centering}p{16pt}<{\centering}|p{11pt}<{\centering}p{11pt}<{\centering}p{16pt}<{\centering}}
\hline
&\multirow{2}{0.5cm}{SyncBN} & \multirow{2}{0.5cm}{GN} & & SN& &  & SyncSN& \\
&        &    &  (8,2) & (8,4) & (8,32)&(8,2)& (8,4) & (8,32) \\
\hline
mIoU$_{\mathrm{ss}}$ &76.3 & 72.6 & 75.9 & \textbf{77.1} & 76.2 &76.0 & \textbf{77.1} & 76.7  \\
mIoU$_{\mathrm{ms}}$ &76.9 & 74.3 & 76.4 & 77.9 & 77.3 &76.6 & \textbf{78.3} & 77.8  \\
\hline
\end{tabular}
\caption{\small{\textbf{Results in Cityscapes validation set by using
PSPNet}  with ResNet50 as backbone network. `$\mathrm{ss}$' and `$\mathrm{ms}$' indicate single-scale and multi-scale inference, respectively.
}}\label{tab:segmentation2}
\end{table}

\subsubsection{Performance Comparisons}
\label{sec:SegComparison}

%
Table~\ref{tab:segmentation} reports mIoU on the ADE20K validation set and Cityscapes test set, by using both single-scale and multi-scale testing. In SN, BN is not synchronized across GPUs, while it is synchronized in SyncSN.
In ADE20K, the best model of SN outperforms SyncBN with a large margin in both testing schemes (38.7 \vs 36.4 and 39.2 \vs 37.7), and improve GN ( \ie channels in each group is $32$ ) by 3.0 and 2.9.
By using SyncSN, the best performance has been further improved to 40.0 for single-scale and 40.4 for multi-scale test.
In Cityscapes,
SN also performs better than SyncBN and GN. For example, the best model of SN surpasses SyncBN by 2.5 and 2.8 in both testing scales.
SyncSN further improves the margins to 7.0 and 4.5.
We see that GN performs worse than SyncBN in these two benchmarks.
%
Note that SyncSN and SN share the same pre-trained models for this task.
It means that we only use SyncSN to finetune the model on both of above datasets.
Fig.\ref{fig:segm_block} compares the importance weights of SN $(8,2)$ in ResNet50 trained on both ADE20K and Cityscapes, showing that different datasets would choose different normalizers when the models and tasks are the same.


\textbf{Finetuning appropriate pretrained models.}
For semantic image segmentation tasks, we observe that ratios pretrained with different batch sizes bring different impacts in finetuning.
Using models that are pretrained and finetuned with comparable batch size would be a best practice for good results.
Otherwise, performance may degenerate.

\begin{figure*}[ht]
\centering
\subfigure[\small{$\lambda_z^\mu$ in \textbf{SN(8,2)} for ADE20K.}] {
\begin{minipage}[c]{\linewidth}
\vspace{-8pt}
\centering
\includegraphics[width=\textwidth,height=30mm]{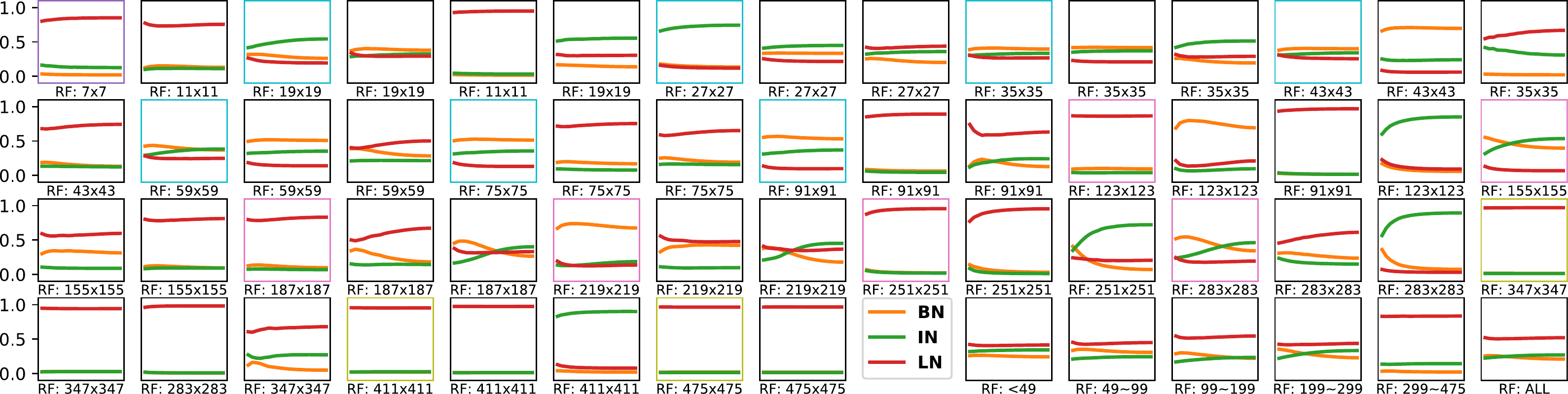}
\end{minipage}%
}%
\hspace{-0.0in}
\subfigure[\small{$\lambda_z^\sigma$ in \textbf{SN(8,2)} for ADE20K.}] {
\begin{minipage}[c]{\linewidth}
\vspace{-8pt}
\centering
\includegraphics[width=\textwidth,height=30mm]{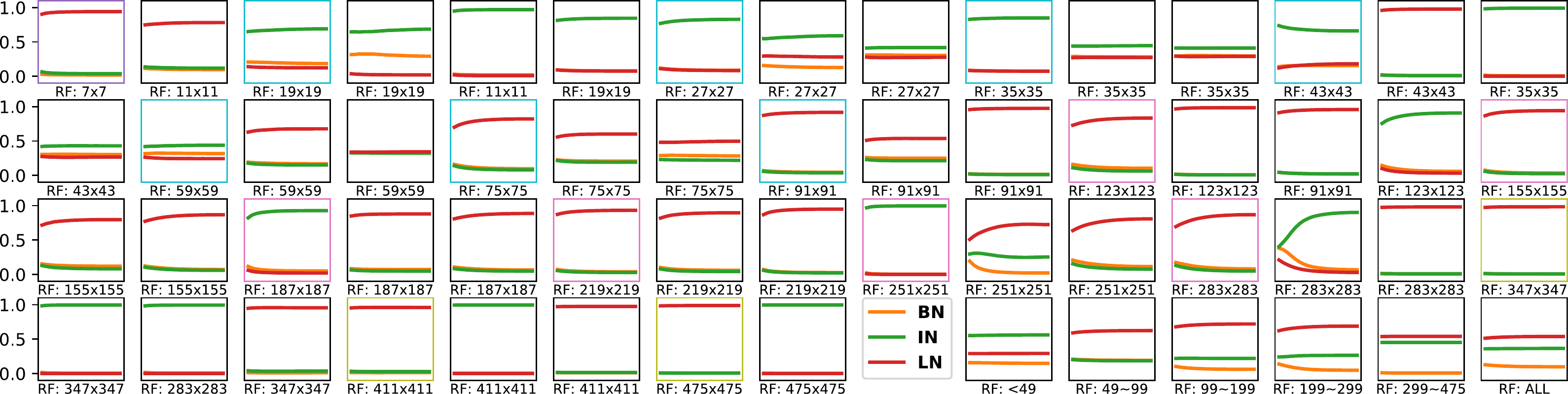}
\end{minipage}%
}%
\hspace{-0.0in}
\subfigure[\small{$\lambda_z^\mu$ in \textbf{SN(8,32)} for ADE20K.}] {
\begin{minipage}[c]{\linewidth}
\vspace{-8pt}
\centering
\includegraphics[width=\textwidth,height=30mm]{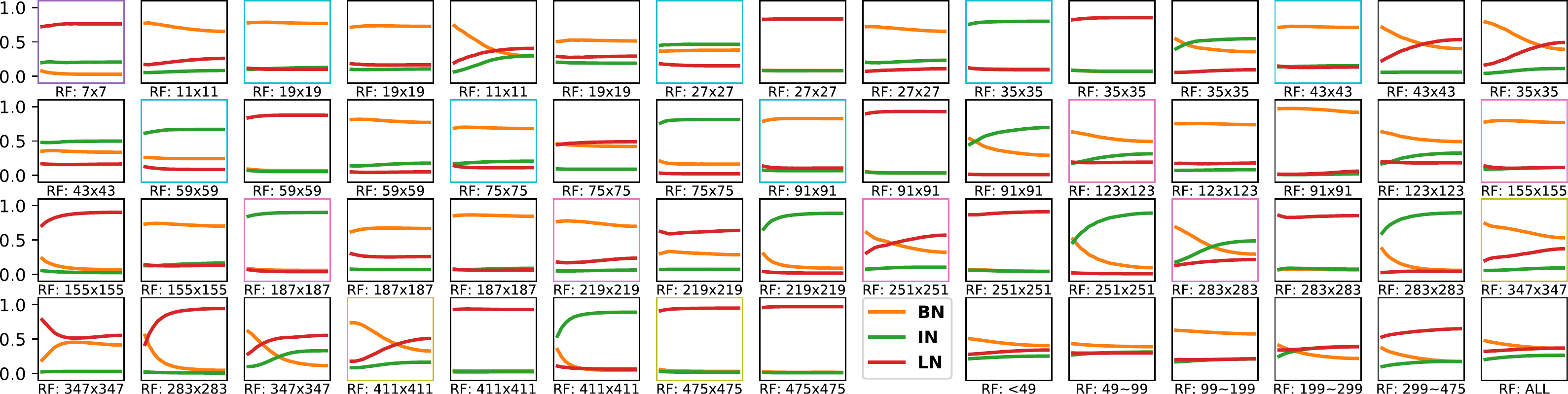}
\end{minipage}%
}%
\hspace{-0.0in}
\subfigure[\small{$\lambda_z^\sigma$ in \textbf{SN(8,32)} for ADE20K.}] {
\begin{minipage}[c]{\linewidth}
\vspace{-8pt}
\centering
\includegraphics[width=\textwidth,height=30mm]{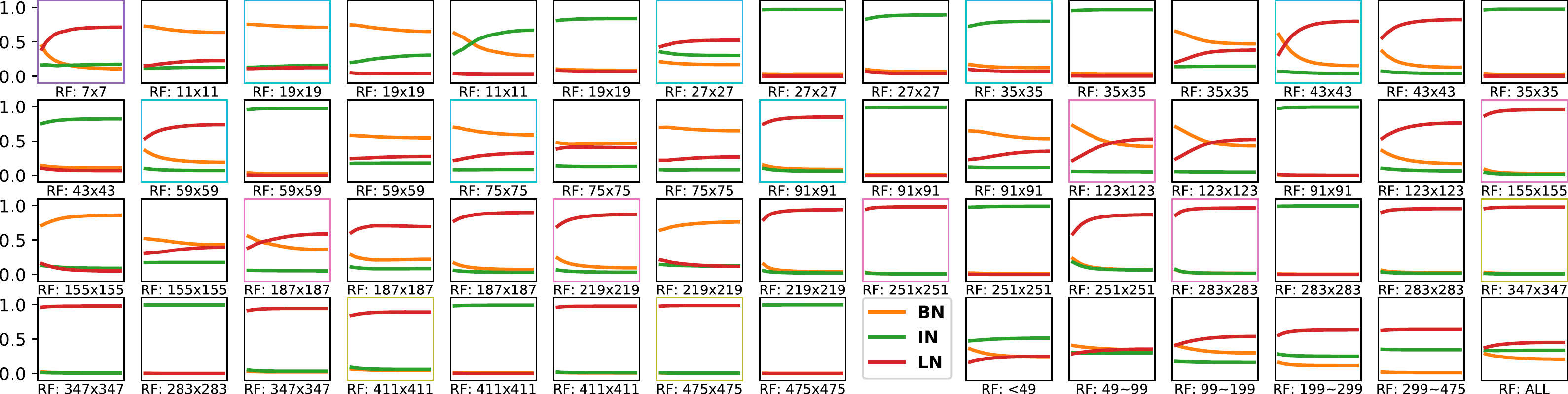}
\end{minipage}%
}%
\vspace{-5pt}
\caption{{Finetuning ResNet50+SN in ADE20K.
}}
\label{appfig:seg_training_ade}
\end{figure*}

In ADE20K, SN$(8,2)$ performs significantly better than the others.
%
In this dataset, SN$(8,32)$ and SN$(8,4)$ may reduce performance, because BN has large ratio in these models, implying that finetuning SN pretrained with large batch to small batch would be unstable.
%
Fig.\ref{appfig:seg_training_ade} shows the ratios when SN$(8,32)$ is used in pretraining but SN$(8,2)$ in finetuning. In line with expectation, the BN ratios are suppressed during finetuning, as the batch statistics become unstable.
%
%
However, these ratios are still suboptimal until training converged, where the BN ratios finetuned from SN$(8,32)$ are still larger than those directly finetuned from SN$(8,2)$, reducing performance in ADE20K.
When adopting SyncSN, the actual ``batchsize" is 16 during training.
SyncSN$(8,32)$ becomes more stable in the finetune phase and achieves best performance under the same setting.

Nevertheless, according to Table~\ref{tab:segmentation}, SN$(8,32)$ and SN$(8,4)$ achieve better results than SN$(8,2)$ in Cityscapes.
This could be attributed to large input image size 713$\times$713 that diminishes noise in the batch statistics of BN (in SN).
Same as ADE20K, SyncSN$(8,32)$ also outperforms the other synchronization models in this dataset.

\begin{figure}
  \centering
  \includegraphics[width=.95\linewidth,height=65mm]{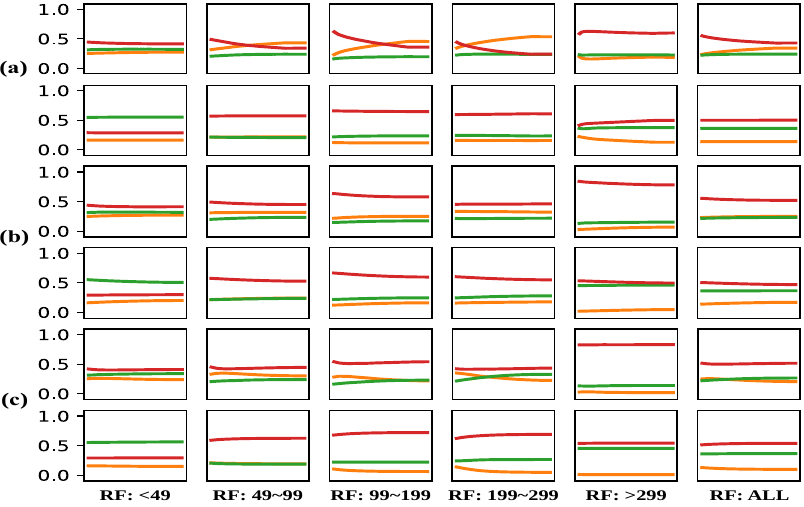}\\
  \caption{{\small \textbf{Ratios for detection and segmentation} including BN ({\color{orange}\textbf{orange}}), IN ({\color{green}\textbf{green}}), and LN ({\color{red}\textbf{red}}). We show $\lambda_z^\mu$ and $\lambda_z^\sigma$ in ResNet50+SN(8,2) finetuned to (a) COCO, (b) Cityscapes, and (c) ADE20K. \vspace{-8pt}
  }}\label{fig:SN_det_seg}
\end{figure}

\textbf{Analysis of Dynamics of Ratios.}
Fig.\ref{fig:SN_det_seg} visualizes ratios in finetuning for semantic image segmentation by using SN$(8,2)$, which are more smooth than pretraining as compared to Fig.\ref{fig:SN8-32-mu-sig}. Intuitively, this is because a small learning rate is typically used in finetuning.
%
In Fig.\ref{fig:SN_det_seg}, IN and LN ratios are generally larger than BN because of small batch size. The lower layers (RF$<$49) prefer IN more than the upper layers (RF$>$299) that choose LN.
In addition, when comparing with detection in COCO, Fig.\ref{fig:SN_det_seg}(b,c) in segmentation have analogue dynamics where BN ratios are gently decreased ($<$0.3).

\textbf{Extension to PSPNet.}
To further evaluate the performance of SN by using complex context modeling, we report the performance of PSPNet~\cite{C:PSPNet} adopting SN on Cityscapes validation set in Table~\ref{tab:segmentation2}.
Both SN and SyncSN outperform GN with a margin.
When compared with SyncBN, the mIoU of the best SN model (\ie SN$(8,4)$) is higher than SyncBN in both inference schemes (77.1 \vs 76.3 and 77.9 \vs 76.9).
However, the ratio of improvement is less than using DeepLab as the backbone model.
On one hand, this could be attributed to PSPNet providing a superior baseline compared with DeepLab.
On the other hand, the spatial pyramid pooling (\ie one type of multi-scale global pooling) may make IN and LN unstable after pooling operation.
By using SyncSN, our best model (\ie SyncSN$(8,4)$) outperforms SyncBN by $0.8$ and $1.4$ in single and multiple scale test.

\begin{figure}[th!]
\begin{center}
   \includegraphics[width=1.0\linewidth]{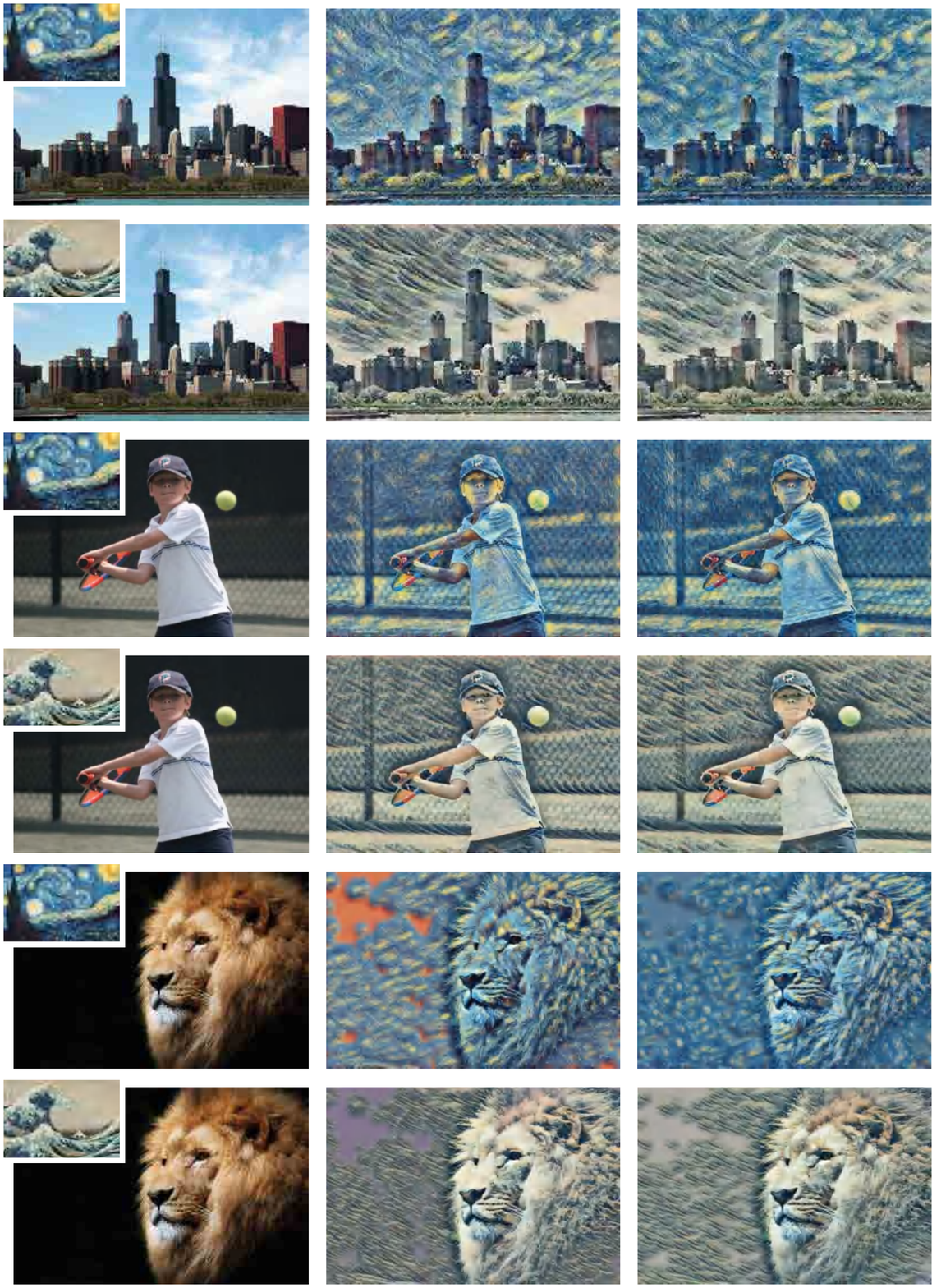}
\end{center}
   \caption{\small{\textbf{Results of Image Stylization.} The first column visualizes the content and the style images. The second and third columns are the results of IN and SN respectively. SN works comparably well with IN in this task.
   }}
\label{fig:vis-handwriting}
\end{figure}

\begin{figure}[h!] \centering
\subfigure[Image Style Transfer] {
\begin{minipage}[c]{0.48\linewidth}
\centering
\includegraphics[width=1\textwidth]{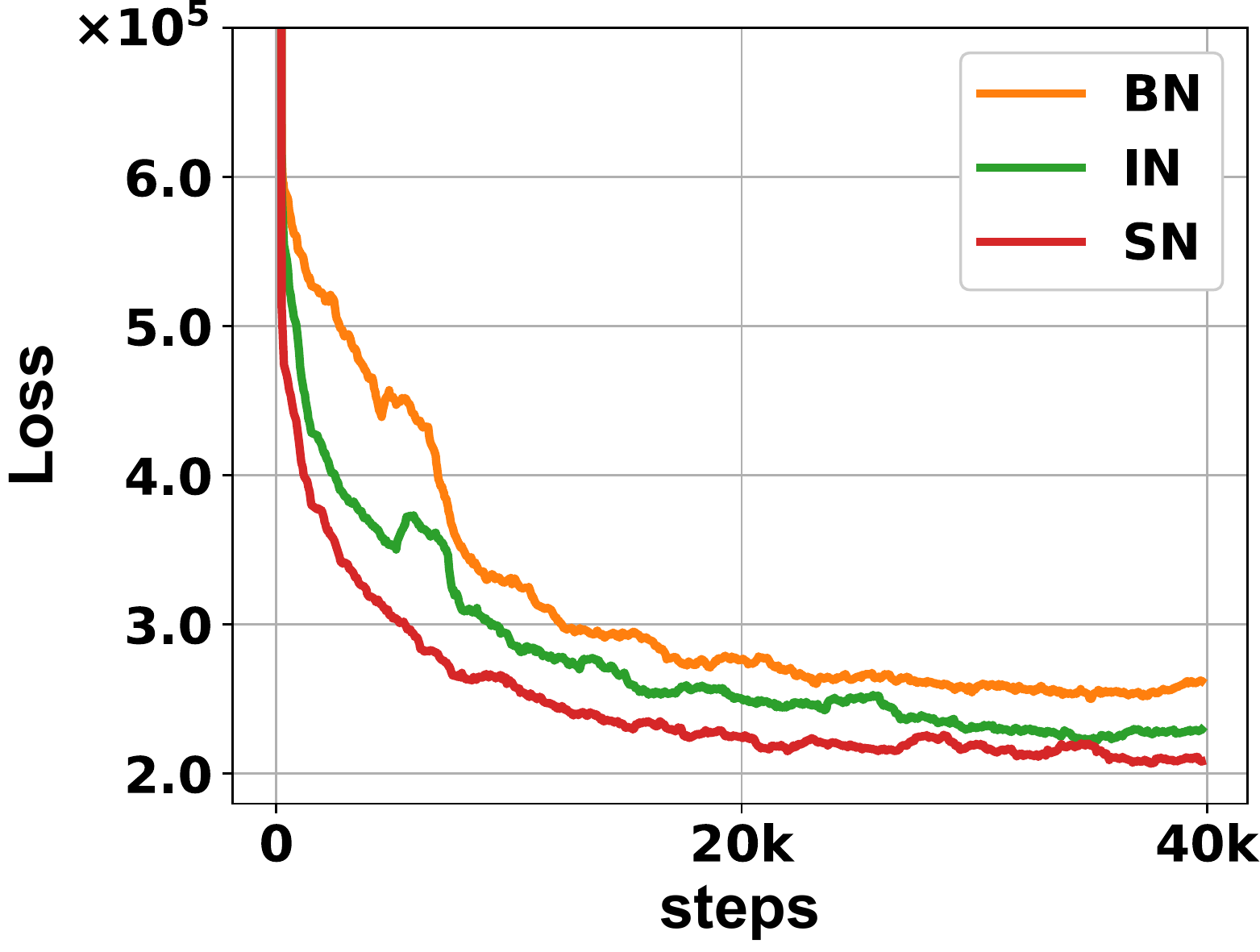}
\end{minipage}%
}%
\hspace{-0.0in}
\subfigure[ENAS on CIFAR-10] {
\begin{minipage}[c]{0.48\linewidth}
\centering
\includegraphics[width=1\textwidth]{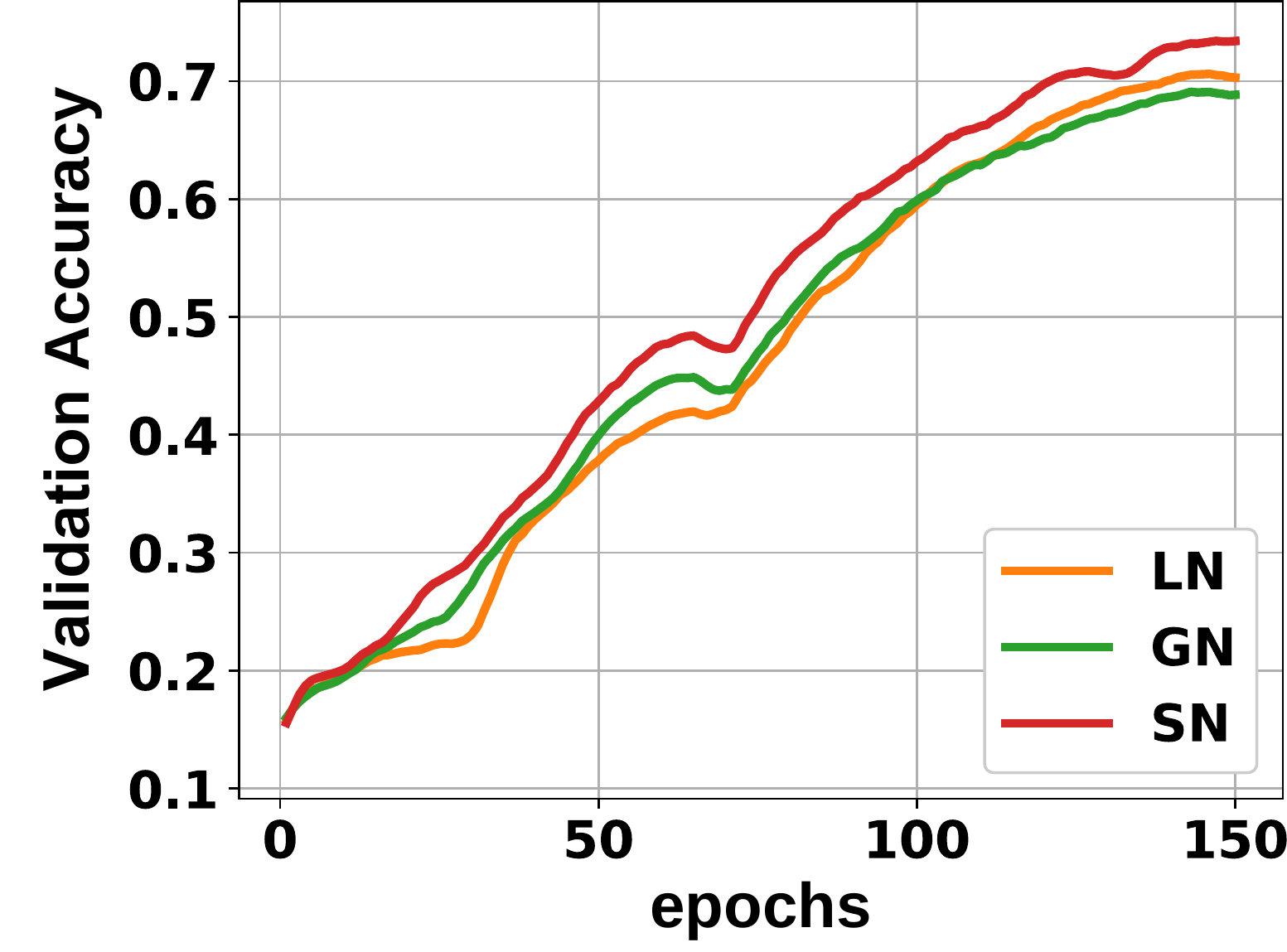}
\end{minipage}%
}%
\caption{\small{(a) shows the losses of BN, IN, and SN in the task of image stylization. SN converges faster than IN and BN. As shown in Fig.\ref{fig:intro} and the supplementary material, SN adapts its importance weight to IN while producing comparable stylization results. (b) plots the accuracy on the validation set of CIFAR-10 when searching network architectures.}}
\label{fig:handwritten-style}
\end{figure}

\subsection{Face Recognition}

We investigate face recognition task in MegaFace dataset~\cite{C:megaface}. We use MS1MV2 dataset~\cite{A:arcface} as our training data.
For data processing, we follow~\cite{A:arcface,C:Sphereface,C:cosface} to generate the normalised face crop (\ie $112 \time 112$) by using five facial point.
We adopt widely used CNN architecture, \ie ResNet50, ResNet100~\cite{C:resnet}, as the backbone networks, and employ ArcFace~\cite{A:arcface} as the loss function.
Following~\cite{A:arcface}, the feature scale is set as $64$ and the angular margin of  ArcFace is $0.5$.
We set the batch size to $64$ for each GPU and train models on $8$ GPUs.
The learning rate is initialized as $0.1$ and divided by $10$ at $[12, 15, 18]$ epoch.
In training phase, we set momentum to $0.9$ and weight decay to $5\times 10^{-4}$.
For both SN and SyncSN, the pretrained model is trained on ImageNet with batch size $32$ per GPU.
During testing, we follow~\cite{A:arcface,C:cosface} to keep the backbone network without the fully connected layer and extract the $512$-$d$ feature for each normalized face.
We employ overlap list\footnote{\url{https://github.com/deepinsight/insightface/tree/master/src/megaface}}
to do dataset filtering to avoid the overlap ID in probe and gallery set,
and then randomly select $1$ million in remaining $1,026,351$ images to evaluate the performance of proposed SN.

Table~\ref{tab:face} shows the performance of SN and SyncSN compared with BN and SyncBN.
When using ResNet50 as the backbone architecture, proposed SN works better than BN no matter exploring synchronization version or not.
For example, the verification result of SN is better than BN and SyncBN by $0.3$ and $0.4$.
SyncSN further improves the margin to $0.5$ and $0.6$.
By using ResNet101, SyncSN still achieves the best verification accuracy, and outperforms BN and SyncBN by $0.2$ and $0.3$, respectively.



\begin{table}[t]
\centering
\begin{tabular}{ c |cc|cc}
\hline
          & BN & SyncBN  & SN  & SyncSN\\
\hline
ResNet50   & 95.3 & 95.2 & 95.6 &  \textbf{95.8} \\
ResNet101  & 96.2 & 96.1 & 95.9 & \textbf{96.4}  \\
\hline
\end{tabular}
\caption{\small{ \textbf{Verification Results of MegaFace dataset} by using different backbone architecture.
}}\label{tab:face}
\end{table}

\begin{table}[t]
\centering
\begin{tabular}{ c |p{30pt}<{\centering}p{30pt}<{\centering}| p{30pt}<{\centering}p{30pt}<{\centering}}
\hline
           & \multicolumn{2}{c|}{  batch=8,~~~length=32 }  & \multicolumn{2}{c}{  batch=4,~~~length=32}  \\
          & top1 & top5  & top1  & top5\\
\hline
     BN   & 73.2 & 90.9 & 72.1 & 90.0\\
     GN        & 73.0 & 90.6  & 72.8 & 90.6\\
     SN        & 73.5 & 91.2 & \textbf{73.3} & \textbf{91.2}\\
\hline
\end{tabular}
\caption{\small{ \textbf{Results of Kinetics dataset.} In training, the clip length of 32 frames is regularly sampled with a frame interval of 2. We study a batch size of 8 or 4 clips per GPU. BN is not synchronized across GPUs.
}}\label{tab:video}
\end{table}

\subsection{Video Recognition}

We evaluate video recognition in Kinetics dataset \cite{kay2017kinetics}, which has 400 action categories.
We experiment with Inflated 3D (I3D) convolutional networks \cite{I3D} and employ the ResNet50 I3D baseline as described in \cite{GN}. The models are pre-trained from ImageNet. For all normalizers, we extend the normalization from over $(H, W)$ to over $(T, H, W)$, where $T$ is the temporal axis.
We train in the training set and evaluate in the validation set. The top1 and top5 classification accuracy are reported by using standard 10-clip testing that averages softmax scores from 10 clips sampled regularly.

Table \ref{tab:video} shows that SN works better than BN and GN in both batch sizes. For example, when batch size is 4, top1 accuracy of SN is better than BN and GN by 1.2\% and 0.5\%. It is seen that SN already surpasses BN and GN with batch size of 8.

\subsection{Artistic Image Stylization}\label{sec:style}

We also evaluate SN in the tasks of artistic image stylization \cite{style}.
We adopt a recent advanced approach \cite{style}, which jointly minimizes two loss functions. Specifically, one is a feature reconstruction loss that penalizes an output image when its content is deviated from a target image,
and the other is a style reconstruction loss that penalizes differences in style (\eg color, texture, exact boundary).
\cite{style,AIN} show that IN works better than BN in this task.

We compare SN with IN and BN using VGG16 \cite{A:VGG} as backbone network.
All models are trained on the COCO dataset \cite{lin2014microsoft}. For each model in training, we resize each image to 256$\times$256 and train for $40,000$ iterations with a batch size setting of $(1,4)$.
We do not employ weight decay or dropout.
The other training protocols are the same as \cite{style}.
In test, we evaluate the trained models on 512$\times$512 images selected following \cite{style}.
Fig.\ref{fig:handwritten-style} (a) compares the style and feature reconstruction losses.
We see that SN enables faster convergence than both IN and BN.
As shown in Fig.\ref{fig:intro} (a), SN automatically selects IN in image stylization.
%
Some stylization results are visualized in Fig.\ref{fig:vis-handwriting}.

\subsection{Neural Architecture Search}\label{sec:nas}

We investigate SN in LSTM for efficient neural architecture search (ENAS) \cite{ENAS}, which is designed to search the structures of convolutional cells. In ENAS, a convolutional neural network (CNN) is constructed by stacking multiple convolutional cells.
It consists of two steps, training controllers and training child models. A controller is a LSTM whose parameters are trained by using the REINFORCE \cite{reinforcement} algorithm to sample a cell architecture, while a child model is a CNN that stacks many sampled cell architectures and its parameters are trained by back-propagation with SGD.
In \cite{ENAS}, the LSTM controller is learned to produce an architecture with high reward, which is the classification accuracy on the validation set of CIFAR-10 \cite{CIFAR10}.
Higher accuracy indicates the controller produces better architecture.

We compare SN with LN and GN by using them in the LSTM controller to improve architecture search.
As BN only achieves limited success in recurrent architecture~\cite{C:BNinRNN} and IN is not applicable to LSTM when it is adopted to process non-image data (\textit{i.e.} the statistics of IN are computed across height and width, which do not exist for non-image data),
%
SN only combines LN and GN in this experiment.
Fig.\ref{fig:handwritten-style} (b) shows the validation accuracy of CIFAR10. We see that SN obtains better accuracy than both LN and GN.


\begin{table}[t]
\small
  \centering
  \caption{{\small Summary of practices that help training CNNs with SN.}}
  \begin{tabular}{p{3pt}p{210pt}}
  \hline
  1. & Initializing ratios of normalizers uniformly \eg $1/3$. Carefully tuning the initial ratios may harm generalization. \\
  2. & Adding dropout with a small ratio (\eg 0.1$\sim$0.2) after each SN layer provides minor improvement of generalization in ImageNet. It reduces over-fitting.\\
  3. & Adding $0.5$ dropout in the last fully-connected layer helps generalization in ImageNet.\\
  4. & A model in pretraining and finetuning should have comparable batch size.\\
  5. & Do not put SN after global pooling when feature map size is 1$\times$1, because IN and LN are unstable after global pooling.\\
  6. & SN with hard ratios improves SN in ImageNet.\\
  7. & SN with hard ratios reduces computational runtime in inference compared to SN. 50\% number of layers in sparse SN select BN for both $\mu$ and $\sigma$, meaning that these BN layers can be turned into linear transformation to reduce runtime in inference.\\
  8. & Synchronizing BN in SN improves generalization.\\
  \hline
  \end{tabular}
  \label{tab:error}
\end{table}


\section{Discussions and Future Work}\label{sec:conclusion}

This work presented Switchable Normalization (SN) to learn different operations in different normalization layers of a deep network.
This novel perspective opens up new direction in many research fields that employ deep learning, such as CV, ML, NLP, Robotics, and Medical Imaging. This work has demonstrated SN in multiple tasks of CV such as recognition, detection, segmentation, image stylization, and neural architecture search, where SN outperforms previous normalizers without bells and whistles.
Our analyses suggest that SN has an appealing characteristic to balance learning and generalization when training deep networks. Investigating SN facilitates the understanding of normalization approaches.
%
To make better use of the proposed SN, we also summarize several important practices when learning to select normalizers in Table~\ref{tab:error}.

Future work will explore SN in more research fields as mentioned above.
Moreover, constructing a theoretical framework to exploring the convergence and generalization ability of SN have importance value and will be treated as future work either.
As an extension of SN, it is valuable to design the algorithm to lean arbitrary normalization operations for different convolutional layers in a deep ConvNet.
%

\bibliographystyle{IEEEtran}
\bibliography{IEEEtran}

\appendices
\section{Back-Propagation of SN}\label{app:BP}


In practice, the back-propagation (BP) stage can be computed by auto differentiation (AD).
For the software without AD, we provide the backward computations of SN for single GPU and multiple GPUs as below.

\textbf{Backward for Single GPU}.
Let $\hat{h}$ be the output of the SN layer represented by a 4D tensor $(N,C,H,W)$ with index $n,c,i,j$.
Let $\hat{h}=\gamma\tilde{h}+\beta$ and $\tilde{h}=\frac{h-\mu}{\sqrt{\sigma^2+\epsilon}}$, where $\mu=w_{\mathrm{bn}}\mu_{\mathrm{bn}}+w_{\mathrm{in}}\mu_{\mathrm{in}}
+w_{\mathrm{ln}}\mu_{\mathrm{ln}}$,
$\sigma^2=w_{\mathrm{bn}}\sigma^2_{\mathrm{bn}}+w_{\mathrm{in}}\sigma^2_{\mathrm{in}}
+w_{\mathrm{ln}}\sigma^2_{\mathrm{ln}}$, and $w_{\mathrm{bn}}+w_{\mathrm{in}}+w_{\mathrm{ln}}=1$.
Note that the importance weights are shared among the means and variances for clarity of notations.
Suppose that each one of $\{\mu,\mu_{\mathrm{bn}},\mu_{\mathrm{in}},\mu_{\mathrm{ln}},\sigma^2,\sigma^2_{\mathrm{bn}},
\sigma^2_{\mathrm{in}},\sigma^2_{\mathrm{ln}}\}$ is reshaped into a vector of $N\times C$ entries,
which are the same as the dimension of IN's statistics.
In the following, let $\mathcal{L}$ be the loss function and $\frac{\partial\mathcal{L}}{\partial\mu_n}$ be the gradient with respect to the $n$-th entry of $\mu$.
We have,

\noindent
\begin{eqnarray}
&&\frac{\partial\mathcal{L}}{\partial\tilde{h}_{ncij}} = \frac{\partial\mathcal{L}}{\partial\hat{h}_{ncij}}\cdot\gamma, \\
&&\frac{\partial\mathcal{L}}{\partial\sigma^2} = -\frac{1}{2(\sigma^2+\epsilon)}\sum_{i,j}^{H,W}\frac{\partial\mathcal{L}}{\partial\tilde{h}_{ncij}}\cdot\tilde{h}_{ncij}, \\
&&\frac{\partial\mathcal{L}}{\partial\mu} = -\frac{1}{\sqrt{\sigma^2+\epsilon}}\sum_{i,j}^{H,W} \frac{\partial\mathcal{L}}{\partial\tilde{h}_{ncij}},
\end{eqnarray}

\noindent
\begin{eqnarray}
\frac{\partial\mathcal{L}}{\partial{h}_{ncij}} &&= \frac{\partial\mathcal{L}}{\partial\tilde{h}_{ncij}}\cdot
\frac{1}{\sqrt{\sigma^2+\epsilon}}
\nonumber\\
&&+\Bigg[ \frac{2w_{\mathrm{in}}(h_{ncij}-\mu_{\mathrm{in}})}{HW}\frac{\partial\mathcal{L}}{\partial\sigma^2}
\nonumber\\
&&+\frac{2w_{\mathrm{ln}}(h_{ncij}-\mu_{\mathrm{ln}})}{CHW}\sum_{c=1}^C \frac{\partial\mathcal{L}}{\partial\sigma^2_c}
\nonumber\\
&&+\frac{2w_{\mathrm{bn}}(h_{ncij}-\mu_{\mathrm{bn}})}{NHW}\sum_{n=1}^N \frac{\partial\mathcal{L}}{\partial\sigma^2_n}
\Bigg]
\nonumber\\
&&+
\Bigg[\frac{w_{\mathrm{in}}}{HW}\frac{\partial\mathcal{L}}{\partial\mu}
+ \frac{w_{\mathrm{ln}}}{CHW}\sum_{c=1}^C \frac{\partial\mathcal{L}}{\partial\mu_c}
\nonumber\\
&&+ \frac{w_{\mathrm{bn}}}{NHW}\sum_{n=1}^N \frac{\partial\mathcal{L}}{\partial\mu_n}
\Bigg],
\end{eqnarray}

\noindent The gradients for $\gamma$ and $\beta$ are

\noindent
\begin{eqnarray}
&&\frac{\partial\mathcal{L}}{\partial \gamma}=\sum_{n,c,i,j}^{N,C,H,W}\frac{\partial\mathcal{L}}{\partial \hat{h}_{ncij}}\cdot\tilde{h}_{ncij}, \\
&&\frac{\partial\mathcal{L}}{\partial \beta} = \sum_{n,c,i,j}^{N,C,H,W}\frac{\partial\mathcal{L}}{\partial \hat{h}_{ncij}},
\end{eqnarray}

\noindent
and the gradients for $\lambda_{\mathrm{in}},\lambda_{\mathrm{ln}}$, and $\lambda_{\mathrm{bn}}$ are

\noindent
\begin{eqnarray}
\frac{\partial\mathcal{L}}{\partial \lambda_{\mathrm{in}}} &=& w_{\mathrm{in}}(1-w_{\mathrm{in}})
\sum_{n,c}^{N,C}\big(\frac{\partial\mathcal{L}}{\partial \mu_{nc}}\mu_{\mathrm{in}} + \frac{\partial\mathcal{L}}{\partial \sigma^2_{nc}}\sigma^2_{\mathrm{in}}\big)\nonumber\\
&&- ~w_{\mathrm{in}}w_{\mathrm{ln}}\sum_{n,c}^{N,C}\big(\frac{\partial\mathcal{L}}{\partial \mu_{nc}}\mu_{\mathrm{ln}}+ \frac{\partial\mathcal{L}}{\partial \sigma^2_{nc}}\sigma^2_{\mathrm{ln}}\big)\nonumber\\
&&- ~w_{\mathrm{in}}w_{\mathrm{bn}}\sum_{n,c}^{N,C}\big(\frac{\partial\mathcal{L}}{\partial \mu_{nc}}\mu_{\mathrm{bn}}+ \frac{\partial\mathcal{L}}{\partial \sigma^2_{nc}}\sigma^2_{\mathrm{bn}}\big),
\end{eqnarray}

\noindent
\begin{eqnarray}
\frac{\partial\mathcal{L}}{\partial \lambda_{\mathrm{ln}}} &=& w_{\mathrm{ln}}(1-w_{\mathrm{ln}})
\sum_{n,c}^{N,C}\big(\frac{\partial\mathcal{L}}{\partial \mu_{nc}}\mu_{\mathrm{ln}}+ \frac{\partial\mathcal{L}}{\partial \sigma^2_{nc}}\sigma^2_{\mathrm{ln}}\big)\nonumber\\
&&-~w_{\mathrm{in}}w_{\mathrm{ln}}\sum_{n,c}^{N,C}\big(\frac{\partial\mathcal{L}}{\partial \mu_{nc}} \mu_{\mathrm{in}}+ \frac{\partial\mathcal{L}}{\partial \sigma^2_{nc}}\sigma^2_{\mathrm{in}}\big)\nonumber\\
&&-~w_{\mathrm{ln}}w_{\mathrm{bn}}\sum_{n,c}^{N,C}\big(\frac{\partial\mathcal{L}}{\partial \mu_{nc}} \mu_{\mathrm{bn}}+ \frac{\partial\mathcal{L}}{\partial \sigma^2_{nc}}\sigma^2_{\mathrm{bn}}\big),
\end{eqnarray}

\noindent
\begin{eqnarray}
\frac{\partial\mathcal{L}}{\partial \lambda_{\mathrm{bn}}} &=& w_{\mathrm{bn}}(1-w_{\mathrm{bn}})
\sum_{n,c}^{N,C}\big( \frac{\partial\mathcal{L}}{\partial \mu_{nc}} \mu_{\mathrm{bn}}+  \frac{\partial\mathcal{L}}{\partial \sigma^2_{nc}}\sigma^2_{\mathrm{bn}}\big)\nonumber\\
&&- ~w_{\mathrm{in}}w_{\mathrm{bn}}\sum_{n,c}^{N,C}\big(\frac{\partial\mathcal{L}}{\partial \mu_{nc}}\mu_{\mathrm{in}}+ \frac{\partial\mathcal{L}}{\partial \sigma^2_{nc}}\sigma^2_{\mathrm{in}}\big)\nonumber\\
&&- ~w_{\mathrm{ln}}w_{\mathrm{bn}}\sum_{n,c}^{N,C}\big(\frac{\partial\mathcal{L}}{\partial \mu_{nc}}\mu_{\mathrm{ln}}+ \frac{\partial\mathcal{L}}{\partial \sigma^2_{nc}}\sigma^2_{\mathrm{ln}}\big).
\end{eqnarray}

\textbf{Forward and Backward for Multiple GPUs}.
Considering multi-GPU synchronization, let $p\in\{1,2,...,P\}$ denote the index of GPU,
the mean and variance on $p$-th GPU can be denoted as  $\mu_p = w_{\mathrm{bn}}\mu_{\mathrm{sybn}}+w_{\mathrm{in}}\mu_{\mathrm{in},p} +w_{\mathrm{ln}}\mu_{\mathrm{ln},p}$ and
$\sigma^2_p=w_{\mathrm{bn}}\sigma^2_{\mathrm{sybn}}+w_{\mathrm{in}}\sigma^2_{\mathrm{in},p}
+w_{\mathrm{ln}}\sigma^2_{\mathrm{ln},p}$,
where $\mu_{\mathrm{sybn}}$ and $\sigma^2_{\mathrm{sybn}}$ indicate the synchronized statistics of batch mean and batch variance on multiple GPUs.
Let $\hat{h}_{p,ncij}$ indicate the output of the SN layer on the $p$-th GPU, we have

\noindent
\begin{eqnarray}
&&\hat{h}_{p,ncij} = \gamma\tilde{h}_{p,ncij}+\beta \nonumber\\
&&\tilde{h}_{p,ncij} = \frac{ h_{p,ncij} - \mu_p }{ \sqrt{ \sigma_p^2 +\epsilon} }
\end{eqnarray}

\noindent
Let $\mathcal{L}$ be the loss function, the gradients for $\gamma$ and $\beta$ are,

\noindent
\begin{eqnarray}
&&\frac{\partial\mathcal{L}}{\partial \gamma}=\sum_{p,n,i,j}^{P,N,C,H,W}\frac{\partial\mathcal{L}}{\partial \hat{h}_{p,ncij}}\cdot\tilde{h}_{p,ncij}, \\
&&\frac{\partial\mathcal{L}}{\partial \beta} = \sum_{p,n,i,j}^{P,N,H,W}\frac{\partial\mathcal{L}}{\partial \hat{h}_{p,ncij}},
\end{eqnarray}

\noindent
and the gradients for $\lambda_{\mathrm{in}},\lambda_{\mathrm{ln}}$, and $\lambda_{\mathrm{bn}}$ are

\noindent
\begin{eqnarray}
\frac{\partial\mathcal{L}}{\partial \lambda_{\mathrm{in}}} &=& w_{\mathrm{in}}(1-w_{\mathrm{in}})
\sum_{p,n,c}^{P,N,C}\big(\frac{\partial\mathcal{L}}{\partial \mu_{p,nc}}  \mu_{\mathrm{in},p}+ (\frac{\partial\mathcal{L}}{\partial \sigma^2_{p,nc} }\sigma^2_{\mathrm{in},p}\big)\nonumber\\
&&- ~w_{\mathrm{in}}w_{\mathrm{ln}}\sum_{p,n,c}^{P,N,C}\big( \frac{\partial\mathcal{L}}{\partial \mu_{p,nc}} \mu_{\mathrm{ln},p}+  \frac{\partial\mathcal{L}}{\partial \sigma^2_{p,nc}} \sigma^2_{\mathrm{ln},p}\big)\nonumber\\
&&- ~w_{\mathrm{in}}w_{\mathrm{bn}}\sum_{p,n,c}^{P,N,C}\big( \frac{\partial\mathcal{L}}{\partial \mu_{p,nc}} \mu_{\mathrm{sybn}}+  \frac{\partial\mathcal{L}}{\partial \sigma^2_{p,nc}} \sigma^2_{\mathrm{sybn}}\big),\nonumber\\
&&
\end{eqnarray}

\noindent
\begin{eqnarray}
\frac{\partial\mathcal{L}}{\partial \lambda_{\mathrm{ln}}} &=& w_{\mathrm{ln}}(1-w_{\mathrm{ln}})
\sum_{p,n,c}^{P,N,C}\big((\frac{\partial\mathcal{L}}{\partial \mu_{p,nc}} \mu_{\mathrm{ln},p}+  \frac{\partial\mathcal{L}}{\partial \sigma^2_{p,nc}} \sigma^2_{\mathrm{ln},p}\big)\nonumber\\
&&-~w_{\mathrm{in}}w_{\mathrm{ln}}\sum_{p,n,c}^{P,N,C}\big(\frac{\partial\mathcal{L}}{\partial \mu_{p,nc}} \mu_{\mathrm{in},p}+ \frac{\partial\mathcal{L}}{\partial \sigma^2_{p,nc}} \sigma^2_{\mathrm{in},p}\big)\nonumber\\
&&-~w_{\mathrm{ln}}w_{\mathrm{bn}}\sum_{p,n,c}^{P,N,C}\big((\frac{\partial\mathcal{L}}{\partial \mu_{p,nc}}\mu_{\mathrm{sybn}}+ \frac{\partial\mathcal{L}}{\partial \sigma^2_{p,nc}} \sigma^2_{\mathrm{sybn}}\big),\nonumber\\
&&
\end{eqnarray}

\noindent
\begin{eqnarray}
\frac{\partial\mathcal{L}}{\partial \lambda_{\mathrm{bn}}} &=& w_{\mathrm{bn}}(1-w_{\mathrm{bn}})
\sum_{p,n,c}^{P,N,C}\big( \frac{\partial\mathcal{L}}{\partial \mu_{p,nc}} \mu_{\mathrm{sybn}}\nonumber + \frac{\partial\mathcal{L}}{\partial \sigma^2_{p,nc}}\sigma^2_{\mathrm{sybn}}\big)\nonumber\\
&&- ~w_{\mathrm{in}}w_{\mathrm{bn}}\sum_{p,n,c}^{P,N,C}\big(\frac{\partial\mathcal{L}}{\partial \mu_{p,nc}} \mu_{\mathrm{in},p}+ (\frac{\partial\mathcal{L}}{\partial \sigma^2_{p,nc}} \sigma^2_{\mathrm{in},p}\big)\nonumber\\
&&- ~w_{\mathrm{ln}}w_{\mathrm{bn}}\sum_{p,n,c}^{P,N,C}\big(\frac{\partial\mathcal{L}}{\partial \mu_{p,nc}} \mu_{\mathrm{ln},p}+ (\frac{\partial\mathcal{L}}{\partial \sigma^2_{p,nc}} \sigma^2_{\mathrm{ln},p}\big).\nonumber\\
&&
\end{eqnarray}

\noindent
and the gradients for mean and variance on a specific GPU are

\noindent
\begin{eqnarray}
\frac{\partial\mathcal{L}}{\partial \sigma^2_{p,nc}} &=& -\frac{1}{2\sqrt{ \sigma^2_{p,nc} +\epsilon}} \sum_{i,j}^{H,W}\frac{\partial\mathcal{L}}{\partial \tilde{h}_{p,ncij}} \cdot \tilde{h}_{p,ncij}  \nonumber\\
&& -\frac{\gamma}{2\sqrt{ \sigma^2_{p,nc} +\epsilon}} \sum_{i,j}^{H,W}\frac{\partial\mathcal{L}}{\partial \hat{h}_{p,ncij}} \cdot \tilde{h}_{p,ncij}, \nonumber\\
\frac{\partial\mathcal{L}}{\partial \mu_{p,nc}} &=& -\frac{1}{\sqrt{ \sigma^2_{p,nc} +\epsilon}} \sum_{i,j}^{H,W}\frac{\partial\mathcal{L}}{\partial \tilde{h}_{p,ncij}} \nonumber\\
&& -\frac{\gamma}{\sqrt{ \sigma^2_{p,nc} +\epsilon}} \sum_{i,j}^{H,W}\frac{\partial\mathcal{L}}{\partial \hat{h}_{p,ncij}},
\end{eqnarray}

\noindent
and the gradients respect to the input can be calculated by

\noindent
\begin{eqnarray}
\frac{\partial\mathcal{L}}{\partial h_{p,ncij}} &=& \underbrace{\frac{\partial\mathcal{L}}{\partial \tilde{h}_{p,ncij}} \cdot \frac{\partial \tilde{h}_{p,ncij}}{\partial h_{p,ncij}}  }_{\text{Term1}} + \underbrace{ \sum_{\ddot{p},\ddot{n},\ddot{c}}^{P,N,C} \frac{\partial\mathcal{L}}{\partial \sigma^2_{\ddot{p},\ddot{n}\ddot{c}} }  \cdot \frac{\partial \sigma^2_{\ddot{p},\ddot{n}\ddot{c}} } {\partial h_{p,ncij}}  }_{\text{Term2}} \nonumber\\
&& + \underbrace{  \sum_{\ddot{p},\ddot{n},\ddot{c}}^{P,N,C} \frac{\partial\mathcal{L}}{\partial \mu_{\ddot{p},\ddot{n}\ddot{c}} }  \cdot \frac{\partial \mu_{\ddot{p},\ddot{n}\ddot{c}} } {\partial h_{p,ncij}} }_{\text{Term3}},
\end{eqnarray}

\noindent
and,

\noindent
\begin{eqnarray}
\text{Term1} = \frac{\partial\mathcal{L}}{\partial \hat{h}_{p,ncij}} \cdot \frac{\gamma}{\sqrt{ \sigma^2_{p,nc} +\epsilon}},
\end{eqnarray}

\begin{eqnarray}
\text{Term2} &=& \sum_{\ddot{p},\ddot{n},\ddot{c}}^{P,N,C} \frac{\partial\mathcal{L}}{\partial \sigma^2_{\ddot{p},\ddot{n}\ddot{c}}} \cdot \big( w_{\mathrm{in}} \frac{\partial \sigma^2_{\mathrm{in},p,nc}}{\partial h_{p,ncij}} \delta_{p\ddot{p},n\ddot{n},c\ddot{c}} \nonumber\\
&& + w_{\mathrm{ln}} \frac{\partial \sigma^2_{\mathrm{ln},p,n}}{\partial h_{p,ncij}} \delta_{p\ddot{p},n\ddot{n}}  +  w_{\mathrm{bn}} \frac{\partial \sigma^2_{\mathrm{sybn},c}}{\partial h_{p,ncij}} \delta_{c\ddot{c}}\big) \nonumber\\
&=& w_{\mathrm{in}} \frac{\partial \sigma^2_{\mathrm{in},p,nc}}{h_{p,ncij}} \frac{\partial\mathcal{L}}{\partial \sigma^2_{p,nc}}  + w_{\mathrm{ln}} \frac{\partial \sigma^2_{\mathrm{ln},p,n}}{h_{p,ncij}} \sum_{\ddot{c}}^C\frac{\partial\mathcal{L}}{\partial \sigma^2_{p,n\ddot{c}}} \nonumber\\
&& + w_{\mathrm{bn}} \frac{\partial \sigma^2_{\mathrm{sybn},c}}{h_{p,ncij}}  \sum_{\ddot{p},\ddot{n}}^{P,N} \frac{\partial\mathcal{L}}{\partial \sigma^2_{\ddot{p},\ddot{n}c}}  \nonumber\\
&=& w_{\mathrm{in}} \frac{ 2(h_{p,ncij} - \mu_{\mathrm{in},p,nc}) }{HW} \frac{\partial\mathcal{L}}{\partial \sigma^2_{p,nc}} \nonumber\\
&& + w_{\mathrm{ln}} \frac{ 2(h_{p,ncij} - \mu_{\mathrm{ln},p,n}) }{CHW} \sum_{\ddot{c}}^C \frac{\partial\mathcal{L}}{\partial \sigma^2_{p,n\ddot{c}}} \nonumber\\
&& + w_{\mathrm{bn}} \frac{ 2(h_{p,ncij} - \mu_{\mathrm{sybn},c}) }{PNHW} \sum_{\ddot{p},\ddot{n}}^{P,N} \frac{\partial\mathcal{L}}{\partial \sigma^2_{\ddot{p},\ddot{n}c}} \nonumber\\
\end{eqnarray}

\begin{eqnarray}
\text{Term3} &=& \sum_{\ddot{p},\ddot{n},\ddot{c}}^{P,N,C} \frac{\partial\mathcal{L}}{\partial \mu_{\ddot{p},\ddot{n}\ddot{c}}} \cdot \big( w_{\mathrm{in}} \frac{\partial \mu_{\mathrm{in},p,nc}}{\partial h_{p,ncij}} \delta_{p\ddot{p},n\ddot{n},c\ddot{c}} \nonumber\\
&& + w_{\mathrm{ln}} \frac{\partial \mu_{\mathrm{ln},p,n}}{\partial h_{p,ncij}} \delta_{p\ddot{p},n\ddot{n}}  +  w_{\mathrm{bn}} \frac{\partial \mu_{\mathrm{sybn},c}}{\partial h_{p,ncij}} \delta_{c\ddot{c}}\big) \nonumber\\
&=& w_{\mathrm{in}} \frac{\partial \mu_{\mathrm{in},p,nc}}{h_{p,ncij}} \frac{\partial\mathcal{L}}{\partial \mu_{p,nc}}  + w_{\mathrm{ln}} \frac{\partial \mu_{\mathrm{ln},p,n}}{h_{p,ncij}} \sum_{\ddot{c}}^C\frac{\partial\mathcal{L}}{\partial \mu_{p,n\ddot{c}}} \nonumber\\
&& + w_{\mathrm{bn}} \frac{\partial \mu_{\mathrm{sybn},c}}{h_{p,ncij}}  \sum_{\ddot{p},\ddot{n}}^{P,N} \frac{\partial\mathcal{L}}{\partial \mu_{\ddot{p},\ddot{n}c}}  \nonumber\\
&=& w_{\mathrm{in}} \frac{ 1 }{HW} \frac{\partial\mathcal{L}}{\partial \mu_{p,nc}} + w_{\mathrm{ln}} \frac{ 1 }{CHW} \sum_{\ddot{c}}^C \frac{\partial\mathcal{L}}{\partial \mu_{p,n\ddot{c}}} \nonumber\\
&& + w_{\mathrm{bn}} \frac{ 1 }{PNHW} \sum_{\ddot{p},\ddot{n}}^{P,N} \frac{\partial\mathcal{L}}{\partial \mu_{\ddot{p},\ddot{n}c}}
\end{eqnarray}

\noindent
where $\delta$ indicates a Dirac Delta function that  $\delta_{t\ddot{t}} = 1$ if $t=\ddot{t}$ and $\delta_{t\ddot{t}}=0$ if $t \neq \ddot{t}$.
Since we adopt synchronized statistics of batch mean and batch variance on multiple GPU, the information from the other GPUs also effect the output of specific GPU in the above formulas, we introduce $\ddot{p},\ddot{n}$ and $\ddot{c}$ to help to describe the such interaction between different GPUs.



\begin{IEEEbiography}[{\includegraphics[width=1in,height=1.25in,clip,keepaspectratio]{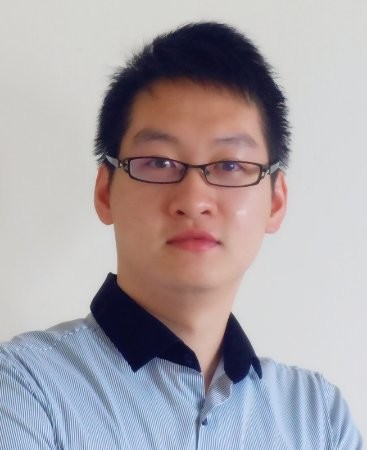}}]{Ping Luo}
is an Assistant Professor in the department of computer science, The University of Hong Kong (HKU). He received his PhD degree in 2014 from Information Engineering, the Chinese University of Hong Kong (CUHK), supervised by Prof. Xiaoou Tang and Prof. Xiaogang Wang. He was a Postdoctoral Fellow in CUHK from 2014 to 2016. He joined SenseTime Research as a Principal Research Scientist from 2017 to 2018. His research interests are machine learning and computer vision. He has published 70+ peer-reviewed articles in top-tier conferences and journals such as TPAMI, IJCV, ICML, ICLR, CVPR, and NIPS. His work has high impact with 7,000 citations according to Google Scholar. He has won a number of competitions and awards such as the first runner up in 2014 ImageNet ILSVRC Challenge, the first place in 2017 DAVIS Challenge on Video Object Segmentation, Gold medal in 2017 Youtube 8M Video Classification Challenge, the first place in 2018 Drivable Area Segmentation Challenge for Autonomous Driving, 2011 HK PhD Fellow Award, and 2013 Microsoft Research Fellow Award (ten PhDs in Asia).
\end{IEEEbiography}
\vspace{-10ex}

\begin{IEEEbiography}[{\includegraphics[width=1in,height=1.25in,clip,keepaspectratio]{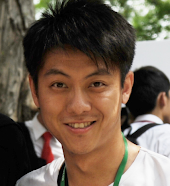}}]{Ruimao Zhang} is currently a Senior Researcher in SenseTime Research. He received the B.E. and Ph.D. degrees from Sun Yat-sen University (SYSU), Guangzhou, China in 2011 and 2016, respectively. From 2017 to 2019, he was a Postdoctoral Research Fellow in the Department of Electronic Engineering, The Chinese University of Hong Kong (CUHK), Hong Kong, China. His research interests include computer vision, deep learning and related multimedia applications. He currently serves as a reviewer of numerous academic journals and conferences, including IJCV, T-NNLS, T-IP, T-CSVT, T-MM, CVPR, ICCV and IJCAI. He is a member of IEEE.
\end{IEEEbiography}
\vspace{-10ex}

\begin{IEEEbiography}[{\includegraphics[width=1in,height=1.25in,clip,keepaspectratio]{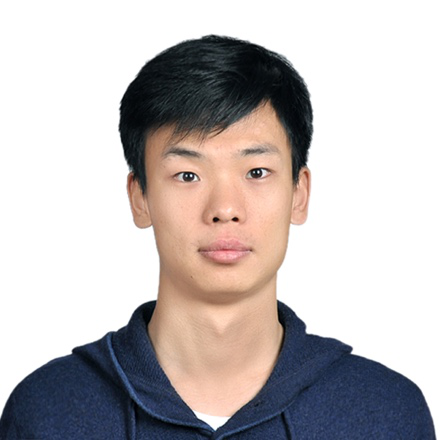}}]{Jiamin Ren} is currently a Researcher in SenseTime Research. He received the B.E. and M.S. degrees from Harbin Institute of Technology(HIT), Harbin, China in 2014 and 2017, respectively.  His research interests include computer vision, machine learning and robotics.
\end{IEEEbiography}
\vspace{-10ex}

\begin{IEEEbiography}[{\includegraphics[width=1in,height=1.25in,clip,keepaspectratio]{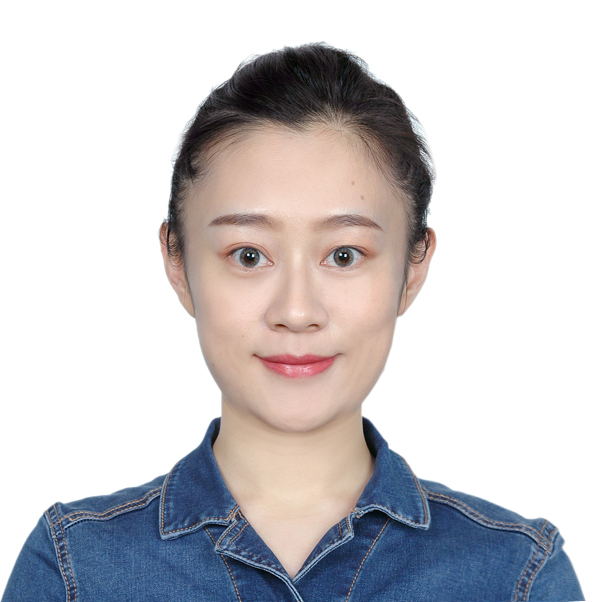}}]{Zhanglin Peng} is currently a Researcher in SenseTime Research. She received the B.E. and M.S. degrees from Sun Yat-sen University (SYSU), Guangzhou, China in 2013 and 2016, respectively. Her research interests include computer vision and deep learning.
\end{IEEEbiography}
\vspace{-10ex}

\begin{IEEEbiography}[{\includegraphics[width=1in,height=1.25in,clip,keepaspectratio]{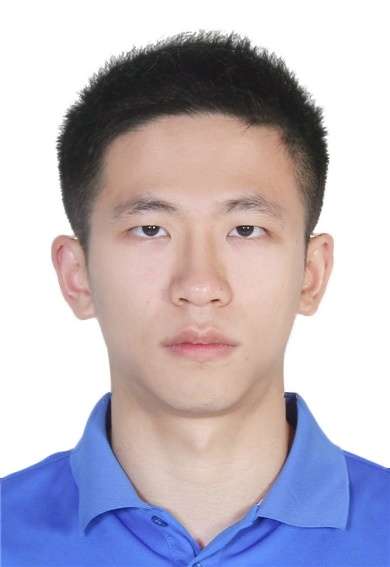}}]{Jingyu Li} received the B.S. degree from The Chinese University of Hong Kong(CUHK), Hong Kong, China. He currently is a Ph.D. student in the Department of Electronic Engineering. In 2017, he joined Sensetime Research as intern where he worked on deep learning for video analysis and normalization algorithm. His research interests include computer vision , deep learning and multimedia data processing. He is a recipient of Hong Kong Postgraduate Fellowship.
\end{IEEEbiography}

\end{document}